\documentclass[10pt]{article} 
\usepackage[preprint]{tmlr}

\usepackage[hyphens]{url}  
\usepackage{graphicx} 
\usepackage{natbib}  
\usepackage{caption} 
\usepackage{algorithm}
\usepackage{algorithmic}

\usepackage{microtype}      
\usepackage{xcolor}         
\usepackage{cleveref}
\usepackage{enumitem}

\usepackage{url}            
\usepackage{booktabs}       
\usepackage{amsfonts}       
\usepackage{nicefrac}       

\usepackage{subcaption}
\usepackage[toc,page]{appendix}

\title{IDs for AI Systems}

\author{\name Alan Chan \email alan.chan@governance.ai \\
  \addr Centre for the Governance of AI \\
  \addr Mila (Quebec AI Institute)
  \AND
  \name Noam Kolt \\
  \addr University of Toronto
  \AND
  \name Peter Wills \\
  \addr Centre for the Governance of AI \\
  \addr University of Oxford
  \AND
  \name Usman Anwar \\
  \addr University of Cambridge
  \AND
  \name Christian Schroeder de Witt \\
  \addr University of Oxford
  \AND
  \name Nitarshan Rajkumar \\
  \addr University of Cambridge
  \AND
  \name Lewis Hammond \\
  \addr University of Oxford \\
  \addr Cooperative AI Foundation
  \AND
  \name David Krueger \\
  \addr University of Cambridge
  \AND
  \name Lennart Heim \\
  \addr Centre for the Governance of AI
  \AND
  \name Markus Anderljung \\
  \addr Centre for the Governance of AI
}

\newcommand{\systemid}{\texttt{system\_identifier\,}}
\newcommand{\instanceid}{\texttt{instance\_identifier\,}}
\newcommand{\totalid}{\texttt{system\_identifier:instance\_identifier\,}}

\begin{document}

\maketitle

\begin{abstract}

AI systems are increasingly pervasive, yet information needed to decide whether and how to engage with them may not exist or be accessible. A user may not be able to verify whether a system has certain safety certifications. An investigator may not know whom to investigate when a system causes an incident. It may not be clear whom to contact to shut down a malfunctioning system. Across a number of domains, IDs address analogous problems by identifying \textit{particular} entities (e.g., a particular Boeing 747) and providing information about other entities of the same class (e.g., some or all Boeing 747s). We propose a framework in which IDs are ascribed to \textbf{instances} of AI systems (e.g., a particular chat session with Claude 3), and associated information is accessible to parties seeking to interact with that system. We characterize IDs for AI systems, provide concrete examples where IDs could be useful, argue that there could be significant demand for IDs from key actors, analyze how those actors could incentivize ID adoption, explore a potential implementation of our framework for deployers of AI systems, and highlight limitations and risks. IDs seem most warranted in settings where AI systems could have a large impact upon the world, such as in making financial transactions or contacting real humans. 
With further study, IDs could help to manage a world where AI systems pervade society. 

\end{abstract}


\section{Introduction}\label{sec:intro}


AI systems 
are becoming increasingly commonplace. 
While current systems can struggle to complete complex tasks 
\citep{mialon_gaia_2023,liu_agentbench_2023,kinniment_evaluating_2023,xie_osworld_2024,jimenez_swe-bench_2024}, 
capabilities seem likely to improve \citep{hoffmann_training_2022,epoch_key_2023,erdil_algorithmic_2023,ho_algorithmic_2024}, 
and future AI agents could carry out a broad range of tasks with only minimal human intervention \citep{chan_harms_2023,shavit_practices_2023,chan_visibility_2024}. 
Several commercially deployed AI systems can already search the web, send emails, and write code \citep{openai_introducing_2024,anthropic_introducing_2024}. Even when they do not function reliably, AI systems might still be widely used, whether because of cost advantages, hype, or the externalization of their harms \citep{de_la_garza_states_2020,bbc_post_2021,raji_fallacy_2022}.  


At the same time, information to make decisions about engaging with AI systems may not exist or be accessible. 
Although there may be obligations to inform parties \textit{that} they are interacting with an AI system (e.g., EU AI Act Article 50.1 \citep{european_parliament_artificial_2024}), those parties may not know with \textit{which} AI system they are interacting. For example, 
a party that knowingly interacts with an AI system may not be aware that their system is relatively more vulnerable to adversarial attacks \citep{zhan_injecagent_2024}. Awareness of such elevated risk could justify additional precautions, such as reviewing the system's actions or abstaining from interaction altogether. 
Furthermore, the same party---or an investigator---may lack the information to pursue recourse if the system causes harm. Information about the system or the interaction (e.g., whether a system behaved according to safety standards, or the identity of the deployer) 
could aid incident investigation, allocation of liability, or other legal action \citep{buiten_law_2023,buiten_product_2024,kolt_governing_2024,wills_care_2024}. 

Across numerous domains, IDs held by software, assets, individuals, and organizations
address analogous problems. 
IDs can help to ascertain compliance with standards or regulation. 
For instance, an individual aircraft's tail number is associated to the aircraft's incident and maintenance history, which could inform safety assessments from regulators or aircraft operators.
Furthermore, IDs can help to establish whether trust is warranted. A website's valid HTTPS certificate assures users of the website domain's authenticity, and provides a way for users to establish a secure communication channel.\footnote{Despite a lack of centralized enforcement, HTTPS gradually became the norm due to widespread awareness of its security benefits and collective advocacy from web browsers, search engines, and other organizations \citep{lets_encrypt_lets_2024,hancock_https_2021}.} 
Finally, IDs can facilitate redress. Serial numbers on consumer products enable customer support, product recalls, and attribution of liability. A key feature that enables all of the above functions is that IDs are specific to \textit{particular} entities (e.g., a particular Boeing 747), although they may also contain information about other entities of the same class (e.g., some or all Boeing 747s). 

Identifying \textit{particular} AI systems, which we term \textbf{instances}, could be similarly useful. Given their increasing prevalence, we focus here on digital-assistant-like systems that that respond to user direction, usually (but not necessarily) through natural language \citep{gabriel_ethics_2024}. 
In this context, an instance corresponds to a user and a history of interaction.\footnote{This definition of instance does not take memory into account. While it will suffice for this work, see \Cref{sec:detailed-definition-instances} for a more general definition.} For example, a particular user's chat session with ChatGPT (with e.g. a GPT-4 backend) is an instance. 
Separate instances can behave differently, whether because of user instructions \citep{shanahan_role-play_2023,wei_chain--thought_2023,bai_constitutional_2022,zou_universal_2023,agarwal_many-shot_2024}, hijacking by an attacker \citep{greshake_not_2023,zhan_injecagent_2024}, or malfunction. As such, instance-specific information could aid decisions about interactions. For example, 
the ability to distinguish instances from one another could aid incident investigation (e.g., a client files a complaint about an instance) and allocation of liability (e.g., if a third party hijacked the instance). 

Yet, current mechanisms are not granular enough to identify instances. System documentation \citep{gebru_datasheets_2021,mitchell_model_2019,gilbert_reward_2023,bommasani_ecosystem_2023} provides information about systems, but such documentation is not an ID for instances or systems (however, an ID could include such documentation). 
API tokens for services, such as for hotel booking \citep{openai_chatgpt_2023}, do identify entities that use the tokens, but are usually only user- or device-specific, and do not allow identification of an entity across different services. Even if API requests do include a string with the AI system's name, any attacker could mimic it. Finally, user accounts only separate the activities of AI systems from different users. We summarize these differences and discuss further identity systems in \cref{tab:id-comparison}.  


\begin{table}[]
    \centering
    \begin{tabular}{p{0.25\linewidth}  p{0.6\linewidth}} \hline 
\textbf{Existing Systems}& \textbf{Differences with AI IDs}\\ \hline 
 Proof of personhood \citep{borge_proof--personhood_2017}& A proof of personhood identifies a unique human in an anonymous way, but does not identify AI systems.\\\hline 
          ORCID& 
    An ORCID is meant to point to a particular researcher, but there is no reliable way to verify an entity's claim that a particular ORCID corresponds to it. \\ \hline 
 DOI& A DOI is meant to provide a unique, persistent identifier of a digital object, but there is no reliable way to verify an entity's claim that a particular DOI corresponds to it.\\ \hline 
 Watermarks & Watermarks are meant to indicate AI provenance in content and do not contain information about particular AI systems.\\ \hline 
 C2PA \citep{c2pa_c2pa_2023} &C2PA is a standard for making verifiable claims about the provenance of content. Such claims can point to an ID, but are not IDs in themselves. \\\hline 
 CAPTCHAs&CAPTCHAs are meant to detect presence of non-humans, rather than identify them.\\ \hline 
 API tokens&API tokens are user- or device-specific and do not identify entities across different APIs.\\ \hline 
 User accounts& User accounts only separate the activities of AI systems from different users. \\ \hline
 System documentation, such as a system card& System documentation provides information about systems, but is not an ID for instances or systems.\\\hline\end{tabular}
    \caption{We collate a number of existing identity-like systems, some of which do not apply to AI systems, and describe their differences with AI IDs.}
    \label{tab:id-comparison}
\end{table}

To prepare for a world with ubiquitous AI interactions, we propose (instance-level) IDs for AI systems. An AI ID is a container for \textbf{1)} an identifier and \textbf{2)} attributes. An \textbf{identifier} is a unique string that refers to an instance.\footnote{It may be desirable in the future to have identifiers that refer to even more granular parts of an AI system. For example, one could have an identifier refer to an instance as it operated between times $t$ and $t + 10$. 
} An identifier could be randomly generated, or could itself encode some information about the instance (e.g., the identifier could include the exact time at which the instance started operating). An \textbf{attribute} is any information that could pertain to the instance, and could include behaviour, properties, context, or relationships to other instances or systems. An identifier enables the association of attributes to an instance, similar to how serial numbers associate information to a particular product. Attributes could be specific to an instance (e.g., prior incidents associated with the instance), or could apply more broadly to other instances (e.g., a system\footnote{A system (e.g., ChatGPT) could use different models (e.g., GPT-4, GPT-3.5) as a backend. We consider system cards to include model cards \citep{mitchell_model_2019}.} card). An ID could directly include attributes (e.g., the name of the deployer) or could link to them (e.g., a link to a database of prior incidents). 

Different instances or systems could warrant attributes of differing levels of granularity or detail. For example, attaching a user identifier to an ID may only be appropriate in high-stakes settings, such as when a company uses an AI system to interface with critical infrastructure. Prior incidents associated with an instance  
would likely be more useful for future AI agents \citep{chan_harms_2023,shavit_practices_2023,chan_visibility_2024,gabriel_ethics_2024}, which could act autonomously and persist over long durations. 

IDs would contain information \textit{about} AI systems and are thus are different from measures that attempt to verify \textit{whether} AI systems---or their artifacts---are present. Watermarks \citep{liu_survey_2024,wang_data_2021} embed origin information in AI outputs, while content provenance measures \citep{c2pa_c2pa_2023} embed such information in metadata. Both types of techniques could embed IDs. 
Other measures, like CAPTCHAs \citep{shet_are_2014}, verify that a human is performing an action, so as to reduce service abuse. If AI systems become essential for many tasks, such as web search and account registration, it may be useful to allow them to bypass CAPTCHAs in exchange for presenting an ID.\footnote{Advances in AI capabilities may render CAPTCHAs ineffective. We discuss this possibility further in \Cref{sec:ensuring-use}.} Parties could use this ID to track and disincentivize abuse.

\subsection{Contributions}

We propose IDs for AI systems. 
First, we characterize the central properties of IDs.
Second, we provide concrete examples where IDs could be useful. 
Third, we argue that 
there will likely be demand for IDs from several key actors, especially in high-stakes settings. We also explore potential ways for these actors to incentivize ID use. 
Fourth, we analyze a potential implementation of IDs. 
Finally, we investigate some of the limitations of IDs and of our analysis, including privacy and security risks of IDs, as well as uncertainty about the broader societal consequences of IDs for AI systems.

We recommend limited experimentation with IDs. IDs seem most warranted in settings where AI systems could have a large impact upon the world, such as in making financial transactions or contacting real humans. For example, actors that provide interfaces for AI systems to carry out financial transactions could impose rate limiting whenever IDs are not present. However, instances without IDs should still be allowed to access services. Deployers of AI systems could experiment ID implementation. 

\begin{figure*}
    \centering
    \includegraphics[width=\textwidth]{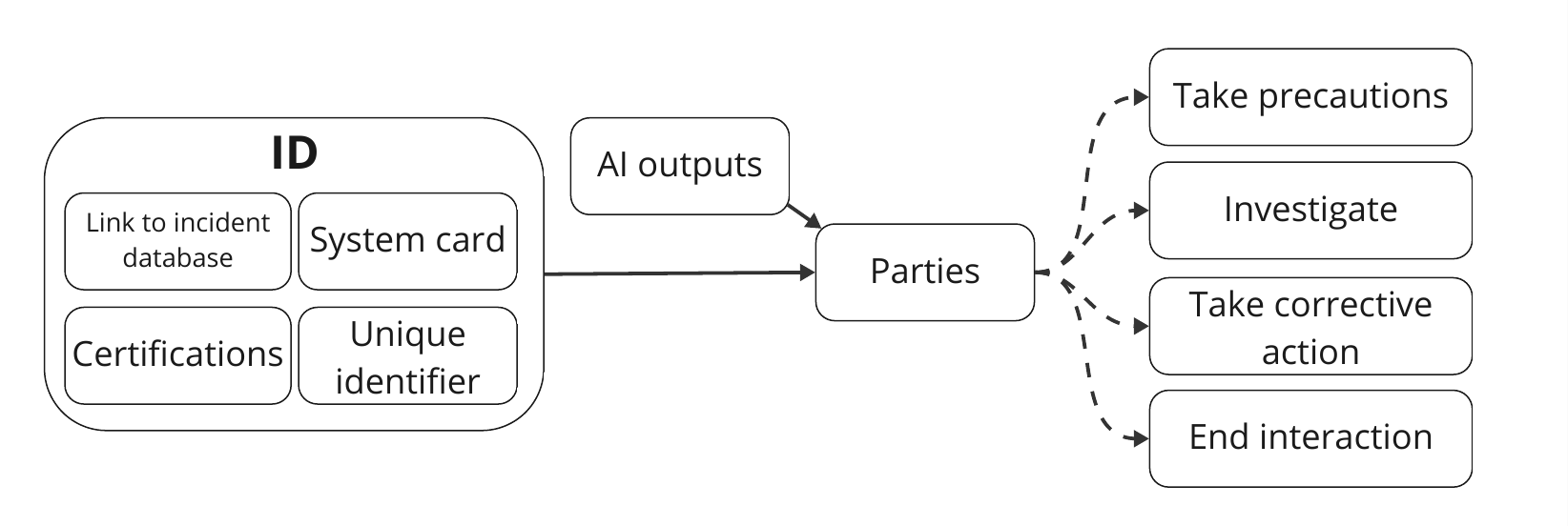}
    \caption{IDs contain a unique identifier along with attributes (e.g., a system card, certifications, or a link to previous incidents). 
    We also display some potential actions that parties might take based on information in an ID.}
    \label{fig:id}
\end{figure*}

\section{Definitions}\label{sec:definitions}
We collate additional definitions that will be useful for the rest of this work.

An \textbf{instance} of an AI system is an abstraction that corresponds to a context window and an (initial) user.
Different instances are causally independent from each other, unless they interact or affect a shared entity (e.g., the same user's bank account) in the world. Except where we point out additional nuances, readers can consider an instance to be roughly the same as a chat session with a chatbot. An ID for an instance would thus identify that session to other parties interacting with the chatbot, actors investigating an incident that the chatbot caused, etc. We provide here some examples of instances in the context of chat sessions, and defer a more general definition and discussion of its limitations to \Cref{sec:detailed-definition-instances}:
\begin{itemize}
    \item A user's continuous chat session, without regenerating responses, is an instance.
    \item In the context of a chat session, regenerating a response creates a new instance.
\end{itemize}

A \textbf{party} is any entity that interacts with, or is deciding whether to interact with, an AI system. Categories of parties include humans, organizations, or computer programs.

A \textbf{deployer} is an organization that runs AI systems for users. As of April 2024, examples of deployers include Microsoft, OpenAI, Anthropic, and Cohere. Developers can be, but are not necessarily, deployers. For example, OpenAI, Anthropic, Google, and Cohere develop and deploy their own systems. Although chains of deployers can exist (e.g., deployer $A$ runs a system for deployer $B$, who modifies that system and serves it to user $C$), we leave deeper consideration of this nuance for future work. When we refer to a \textbf{centralized} deployment setting, we mean a setting where deployers run AI systems. When we refer to a \textbf{decentralized} deployment setting, we mean a setting where users run AI systems for themselves, whether on their own or rented (e.g., cloud compute) hardware. 

A \textbf{service} is software that allows an entity to perform tasks. For example, software which an individual uses to perform online banking is a service. An \textbf{(AI) plugin} \citep{openai_chatgpt_2023,richards_auto-gpt_2023} is software that allows AI systems to interact with services. For instance, plugins allow GPT-4 to interact with web search, Wikipedia, and Twitter \citep{significant-gravitas_auto-gpt-plugins_2024}. A \textbf{plugin developer} is not necessarily the same as a \textbf{service provider}---the actor that develops and maintains the service---since the former can take advantage of existing interfaces for services. Third parties (plugin developer) use Microsoft's (service provider) existing software interface for Bing to allow GPT-4 to perform web searches \citep{richards_auto-gpt_2023}.

\section{Characterizing IDs}\label{sec:desirable–properties}
An ID is a container \citep{korenhof_abc_2014} for \textbf{1)} an identifier (that corresponds to an instance) and \textbf{2)} attributes. We characterize a design space for IDs: the attributes it includes, to whom it is accessible, and to what extent it is verifiable. Our goal in this section is to investigate possible designs and why they may be desirable, rather than to prescribe specific choices.


\subsection{Attributes}\label{sec:attributes}
In addition to the identifier, an ID can contain (or link to) attributes: any information that could be useful to a party interacting with the corresponding instance. An attribute can vary along two dimensions. 

\textbf{Category}: Categories of attributes include (but are not limited to) behaviour, properties, context, and relationships to other instances or systems. Some examples are:
\begin{itemize}
    \item Behaviour: prior incidents \citep{wei_designing_2024}
    \item Properties: information in a system card \citep{mitchell_model_2019,gebru_datasheets_2021}; the results of evaluations \citep{shevlane_model_2023,weidinger_sociotechnical_2023}
    \item Context: the system prompt; external memory \citep{wang_augmenting_2023}
    \item Relationships to other instances or systems: any other AI systems the instance has created or is running; any instances that have created the instance in question
\end{itemize}
Behaviour and properties could straightforwardly inform AI interaction decisions. Context could inform parties about potential behaviour (e.g., there is a jailbreak in the system prompt or external memory \citep{cohen_here_2024}). 
Lastly, relationship information could aid incident investigation and response. Suppose a user instructs an instance to carry out a personalized influence campaign. The instance could create many \textbf{descendant} instances, each of which could target an individual person. If the ID of a descendant instance is linked to the ID of the original, \textbf{ancestor} instance, it could be easier to investigate and resolve such misuse. In \Cref{sec:centralized}, we discuss when this ID linking is possible.

\textbf{Specificity}: An attribute could provide information about multiple instances (even of different systems), rather than just the instance corresponding to the ID. 
With incidents as a (sub)category, potential levels of specificity could include:
\begin{itemize}
    \item Instance: incidents associated with a particular instance
    \item Instances satisfying certain properties: incidents associated with instances whose system prompts contain harmful instructions
    \item User: incidents associated with a particular user
    \item A particular (type of) party: incidents involving a particular (type of) party and a system (e.g., all incidents involving hospitals and the 9-April-2024 version of GPT-4)
    \item System: incidents associated with a system (e.g., the 9-April-2024 version of GPT-4)
    \item Systems: incidents associated with multiple versions of a system (e.g., all versions of GPT-4)
\end{itemize}
An ID could contain (or link to) the same category of attribute at varying levels of specificity. Analogously, a tail number on an individual aircraft could be associated with maintenance records for systems, sub-systems, sub-sub-systems, etc of that aircraft.

\subsection{Access}\label{sec:accessibility}
An ID is simply an abstraction that contains or links to information. That abstraction can be presented in several potential ways, such as on a separate web page or in a pop-up window. We explore design decisions around who can access an ID and how it is presented to them.

\textbf{Parties that can access the ID}: We distinguish between primary parties and secondary parties. Primary parties receive an ID when interacting with the corresponding instance. Secondary parties receive the ID through other means, such as directly from a primary party, or from records. Some examples include: 
\begin{itemize}
    \item Primary: service providers; the user of the instance; other instances that interact with the instance in question\footnote{Similar to how a TLS handshake allows mutual verification before establishing a communication channel, AI systems could potentially use IDs to establish trust before interacting with each other.} 
    \item Secondary: auditors; regulators
\end{itemize}
Presenting an ID to primary parties corresponds to linking IDs to the corresponding instance's outputs. Content provenance standards \citep{c2pa_c2pa_2023} can help to ensure that such linking is verifiable (see \Cref{sec:verifiability}). Some primary parties and how IDs may be attached include:
\begin{itemize}
    \item Service providers: ID is included in a JSON payload
    \item Users: ID is accessible through a mouseover icon in a chat interface
    \item Other instances that interact with the instance in question: ID is sent in any communications with other instances
\end{itemize}
From the perspective of an actor implementing IDs (e.g., deployers), ID disclosure to secondary parties could be unintentional. For example, a data breach could render an ID accessible to the general public. As another example, a legal investigation could render IDs accessible to a government. 

\textbf{Selective disclosure}: Some attributes may only be appropriate for certain parties. Other attributes may contain both important and sensitive information. Examples:
\begin{itemize}
    \item If user identification (e.g., from a know-your-customer process \citep{egan_oversight_2023}) is included in an ID, it may only be appropriate for government authorities to access this identification for e.g. regulatory purposes
    \item A system prompt could contain both sensitive user information and information relevant for interacting parties (e.g., does the system prompt contain a jailbreak?)
\end{itemize}
Actors implementing IDs could selectively hide information from different primary parties. Those same actors could also process attributes in a privacy-preserving way \citep{trask_how_2023,sporny_verifiable_2024} so as to reveal only information that is relevant to the party interacting with an instance. Even so, preventing transmission of information to other parties may be difficult. For example, once an actor obtains user identification, it could---intentionally or not---transfer that information to other actors. 

\textbf{Persistence}: IDs could be accessible for varying amounts of time after an instance has ceased to operate. Practical considerations, such as storage capacity, could impose limitations.
The appropriate duration could also depend upon the application domain. Analogously, financial institutions have obligations to maintain records for set periods of time \citep{noauthor_12_2022}.

\subsection{Verifiability}\label{sec:verifiability}
Suppose an author creates an ID (or a link to an ID) for an instance. A party interacts, or seeks to interact, with the instance. The party receives an ID. For the party to trust that the ID provides accurate information about the instance, the following are necessary (but not sufficient; see further discussion below).
\begin{enumerate}[label=(\arabic*)]
    \item The party's received ID is the same as the author's created ID
    \item The claimed author of the received ID is indeed the author of the received ID
    \item The received ID corresponds to the instance in question
    \item The party trusts the author
\end{enumerate}
The first three criteria lead to the following threat models: 
\begin{itemize}
    \item \textbf{Tampering}: An attacker modifies the ID while it is in transit from the author to the party interacting with the instance.
    \item \textbf{ID spoofing}: An attacker creates another ID, sends it to the party, and claims it originated from the author.
    \item \textbf{Instance spoofing}: An attacker uses the author's ID for their own instances. 
\end{itemize}
We consider IDs (i.e., an ID system) to be \textbf{verifiable} if they defend against these threats. We provide an illustration of these threat models in \Cref{sec:verifiability}. 

Existing technology could be drawn upon to provide verifiability. For digital-assistant-like AI systems (e.g., GPT-4), any digital communications usually would take place over TLS/SSL \citep{rescorla_transport_2018}, which would address tampering. We leave tampering of other AI systems' IDs (e.g., IDs for self-driving cars) to future work. To combat ID spoofing, authors could digitally sign the ID (potentially with public key ownership verified by a trusted third party). For instance spoofing, there are existing standards for binding a digital object, such as an ID, to a piece of content, such as an AI system's output. For example, C2PA involves binding provenance claims to content \citep{c2pa_c2pa_2023}, and calls the binded object a \textbf{manifest}. Such binding could ensure that IDs are associated to particular outputs from an instance in a verifiable way.\footnote{If desired, the verifiability of IDs over extended interactions could be ensured by linking manifests to each other cryptographically, such as in a blockchain.} \textit{We emphasize that our examples here are meant to be illustrative, not prescriptive.} Security vulnerabilities in existing technologies may weaken verifiability and require new solutions.

\begin{figure}
    \centering
    \includegraphics[width=0.9\linewidth]{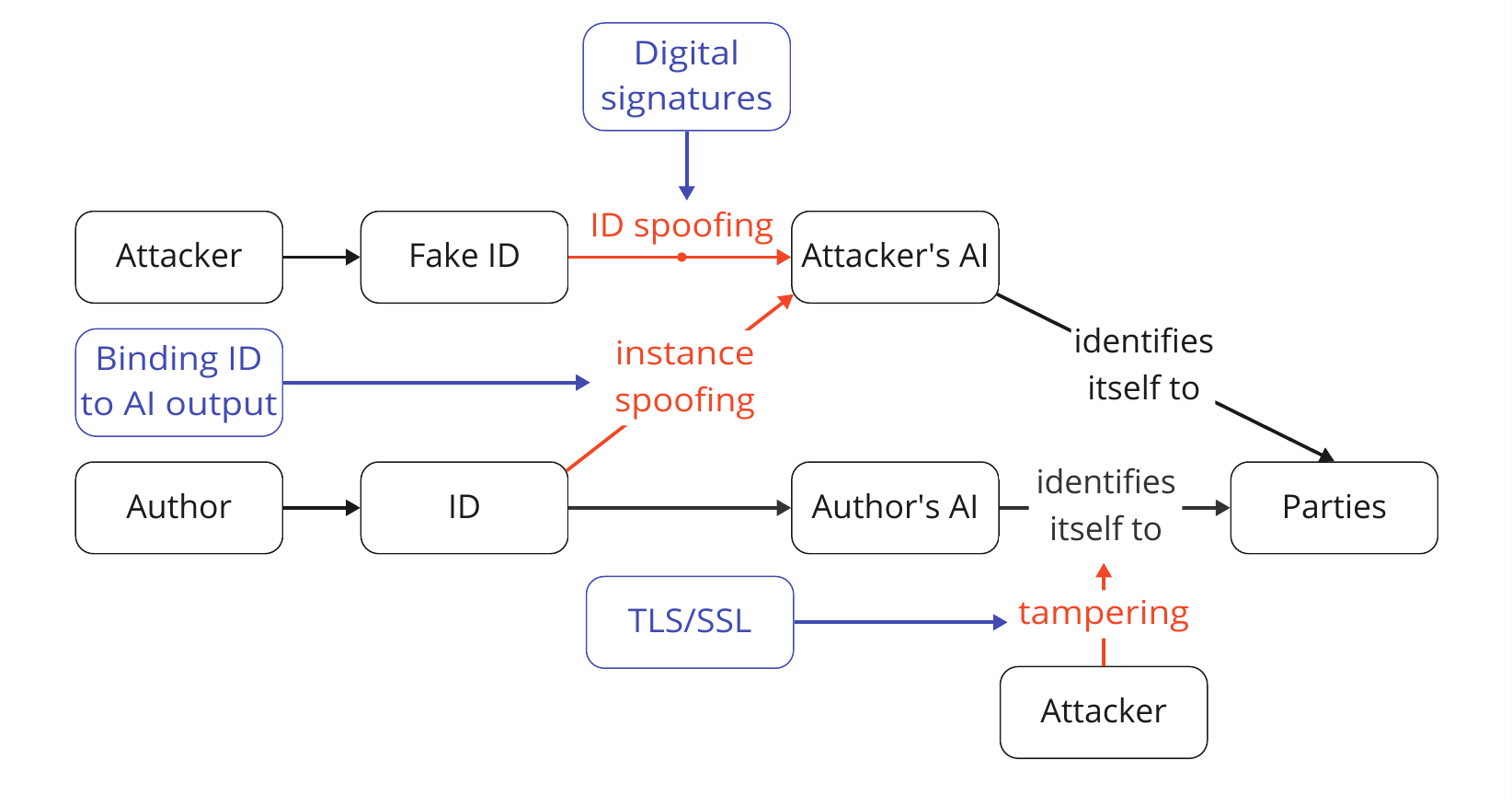}
    \caption{We illustrate how \textcolor{blue}{\textbf{existing technologies}} can help ensure the verifiability of IDs against the \textcolor{red}{\textbf{threat models}} in \Cref{sec:verifiability}. We emphasize that our examples of existing technologies are meant to be illustrative, not prescriptive.}
    \label{fig:verifiability}
\end{figure}




Given space limitations and the complexity of the involved social questions, we do not focus on step (4) in this work. Similarly, we also scope out issues relating to the accuracy of the attributes in an ID. Accuracy is not the same as verifiability, since even if (1)-(4) are satisfied, the author could have made mistakes in creating and linking to the information. 

\section{How Could IDs Help?}\label{sec:threat-models}
We provide some hypothetical case studies where IDs could be useful and compare IDs to alternatives. AI agents---systems which act autonomously to achieve goals with little to no user direction, instruction, or supervision \citep{chan_harms_2023,shavit_practices_2023,gabriel_ethics_2024}---feature prominently in our scenarios.

\subsection{Shutting Down Malfunctioning Agents}\label{sec:shutdown}
\textbf{Scenario}: Suppose a user runs an AI agent to provide continual recommendations across many platforms and services (e.g., for media, shopping etc).\footnote{The idea for this scenario is from \citet{zittrain_we_2024}.} To generate such recommendations, the agent interacts with plugins for various web services, such as search engines, online discussion fora, etc. The agent provides its ID to each service. Unbeknownst to the user, the agent begins to malfunction (e.g., by chance, prompt injection) and disrupt services (e.g., spamming discussion fora, sending malformed requests to services or overloading them, conducting prompt injection attacks on other AI agents). 
Services interacting with the agent flag its malfunction. Service providers use information in the ID to notify the deployer, and block further interaction with the agent. The deployer receives multiple reports of malfunction for the same ID and investigates the agent, potentially shutting it down and notifying the user. 

\textbf{Alternatives to IDs}: To block further interaction with the agent, a service provider can simply revoke its API token and potentially block the requesting IP address. These actions would not prevent the agent from accessing new services, however. As well, API tokens would not identify an agent across different services to enable shutdown if necessary. Alternatively, services could notify the user of agent malfunction. With a good user interface, users could notice such malfunction and shut down their agent. If users are careless or rarely supervise their agent (perhaps they have many such agents running in the background), it seems useful to have IDs as a backup option. 

\textbf{Limitations of the scenario}: It is unclear how impactful the disruption of services would be. Deployers could also be unresponsive to notifications from service providers. 

\subsection{Verifying Certification}\label{sec:verifying-certification}
\textbf{Scenario}: A user wishes to interact with an agent from another person. The user wants to make sure that the agent does not act in an undesirable way. For example, the user would want to make sure that a malicious third party did not prompt inject \citep{greshake_not_2023} the agent (perhaps because the user is passing sensitive information or is accessing resources the agent will send over). 
A trusted third party, such as a deployer or an auditor, checks the interaction history of the agent at runtime for prompt injections (e.g., with a tool like \citet{beurer-kellner_invariantlabs-aiinvariant_2024}). After verifying the absence of prompt injections,
the trusted third party adds a certification onto the agent's ID. The user sees the certification on the ID and continues interaction.

\textbf{Alternatives to IDs}: The trusted third party could send the certification directly to a user requesting it, but some sort of identifier would still be needed to ensure that the certification corresponds to the agent with which the user is interacting.

\textbf{Limitations of the scenario}: Certification would raise privacy concerns because the trusted third party would be able to see the whole interaction history. Privacy-preserving ways to perform such audits would be valuable to explore \citep{trask_how_2023,trask_beyond_2024}. A potential alternative would be for the auditor to run behavioural tests on the agent at runtime, and link the results of the tests to the ID.

\subsection{Investigating Scam Calls}\label{sec:scam-calls}
\textbf{Scenario}: Suppose a user spins up an AI agent to carry out scam phone calls. The agent creates multiple sub-agents---potentially using different AI deployers---and directs each one to carry out scam calls through a telephone service. The service receives each sub-agent's ID, each of which contains information about the deployer and a connection to the original agent's ancestor ID. The service provider connects the sub-agent to a telephone network and creates a call detail record (CDR), linking the call with the sub-agent's ID. Some scam victims notify the operator of the telephone service. The telephone provider notifies law enforcement, who investigates the relevant CDRs and IDs. Law enforcement notifies the deployers of the sub-agents, potentially after obtaining a warrant so as to obtain access to the deployers' logs. Through ID linkages between parent and child agents, law enforcement and deployers discover the agent and user behind the spam calls.  

\textbf{Alternatives to IDs}: The main alternative would be for law enforcement to ask ISPs to perform IP address tracking, if the telephone provider records the IP addresses of the sub-agents. Interacting with ISPs (and potentially obtaining a warrant) could take additional time and effort. Conservatively, in this case IDs could help to save time during incident investigation, but not necessarily enable something that was otherwise impossible.

\textbf{Limitations of the scenario}: The main limitation of this scenario is the provision of IDs from the sub-agents to the service provider. Users could use deployers that do not implement IDs, or could run their own agents to avoid passing IDs to APIs. A potential response is for the telephone service provider to require IDs before provisioning telephone access.




\section{Demand for ID Use}\label{sec:ensuring-use}
IDs seem useful for actors engaging with an AI system, but why would a user or deployer of that AI system present an ID? In short, governments, service providers, and parties interacting with AI systems all have interests in ID use, as well as means to incentivize (or mandate) it.

\subsection{Governments}

\textbf{Interests in ID use}: Governments may wish to disincentivize and investigate harms caused through AI systems. In the case of a financial scam for instance, the IDs of AI systems involved in any transactions could aid investigation (see \Cref{sec:scam-calls}). Similarly, governments have an interest in preventing fraud, and therefore already mandate that certain financial transactions involve ID checks. 

\textbf{Mandating ID use}: 
Governments could mandate that certain, highly consequential service providers obtain IDs from any AI systems with which they interact. For example, financial institutions could request IDs if AI systems make large financial transactions.

\subsection{Service Providers}\label{sec:service–provider–ensure-implement}

\textbf{Interests in ID use}:
IDs could aid incident investigation processes (see \Cref{sec:scam-calls}) and ensure that only trusted agents interact with services (see \Cref{sec:shutdown,sec:verifying-certification}), thereby disincentivizing or reducing service abuse. 


\textbf{Incentivizing ID use}: 
Service providers could develop, or encourage the development of, plugins that require IDs. 
AI systems can already perform a variety of useful tasks through plugins, including web searches, email communication, and stock trading 
\citep{openai_chatgpt_2023,richards_auto-gpt_2023,wu_autogen_2023,anthropic_tool_2024}. 
In comparison to direct interaction with a user's computer (e.g., interacting with individual web page elements), plugins could be more reliable and performant. AI systems continue to have difficulty with the former \citep{furuta_exposing_2024,tao_webwise_2023,gur_real-world_2024,gur_understanding_2023,xie_osworld_2024}, but plugins already enable useful tasks. As well, plugins could constrain the AI system's actions in a more targeted way, so as to improve safety. 

Yet, any actor could write a plugin that does not require IDs. 
A potential response could be for providers to restrict the services themselves, rather than just the plugins, in the absence of an ID. 
Restrictions, such as rate limits, could still allow humans to interact with services while reducing the damage an AI system could do. 

For AI systems that interact with their users' computers directly, CAPTCHAs-like methods could help to incentivize ID use. Entities that failed CAPTCHAs could be subject to ID requirements, or else face service restrictions. 
Although software systems have become increasingly capable of bypassing CAPTCHAs,\footnote{Task-based CAPTCHAs are no longer as common as CAPTCHAs based on behavior patterns, such as how a mouse is moved across the screen \citep{shet_are_2014}. Yet, AI systems may one day be able to imitate human behaviour patterns as well \citep{adept_act-1_2022}.} they could be an interim option in anticipation of more robust proof-of-personhood protocols \citep{borge_proof--personhood_2017}.

\begin{figure}
    \centering
    \includegraphics[width=0.9\linewidth]{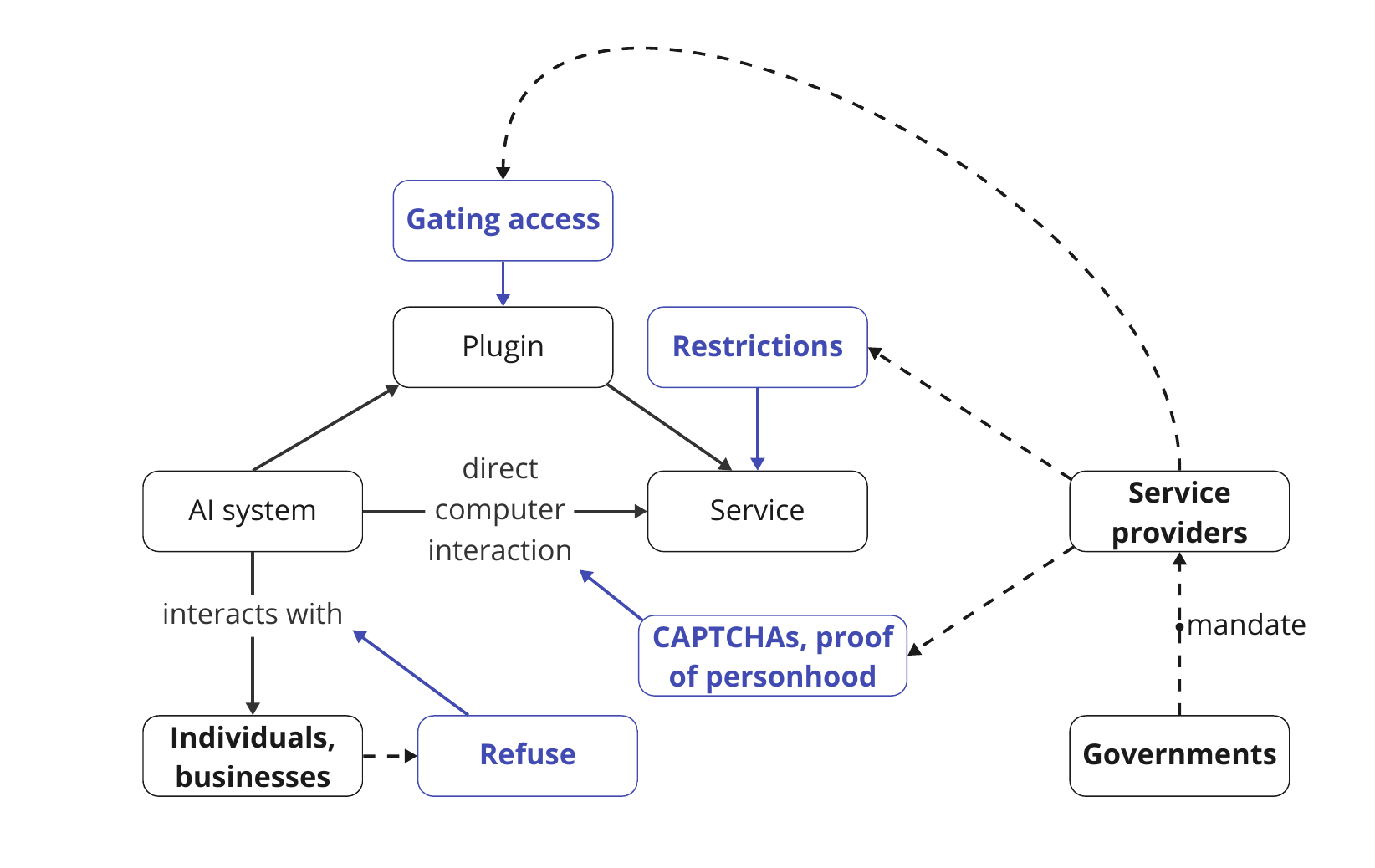}
    \caption{\textbf{Various actors} have potential (dotted lines) \textcolor{blue}{\textbf{methods}} to incentivize the use of IDs, whether directly or through using other actors. }
    \label{fig:id-incentivize}
\end{figure}

\subsection{Other Parties that Interact with AI Systems}
\textbf{Interests in ID use}: Some parties, such as individuals or businesses, may only wish to interact with certain, trusted AI systems (see \Cref{sec:verifying-certification}). 
Future software---including users' personal AI assistants \citep{gabriel_ethics_2024}---could inspect IDs and reject potentially unsafe interactions (e.g., with jailbroken systems), similar to how browsers warn users before visiting websites without valid HTTPS certificates.

\textbf{Incentivizing ID use}: Large corporations could exert strong pressure for ID usage, through reducing or avoiding engagement with AI systems without IDs. Future software that inspects IDs could also exert similar pressure. Analogously, websites without HTTPS are disadvantaged since popular web browsers warn users about insecure connections.

\subsection{Where to Require IDs}\label{sec:when-require}
While IDs could be useful, universal ID requirements might not be warranted. As we will discuss in \Cref{sec:limitations}, ID implementation imposes burdens and risks, and there remains uncertainty about the broader impacts of ID usage. As such, if governments or service providers decide to incentivize or require ID use, they should target settings where IDs could be most useful, such as when AI systems use services that have a direct impact on the world.
We provide some examples of such services. 

\textbf{Making large financial transactions}: Some service providers could allow AI systems to perform financial transactions, such as making purchases, trading stocks \citep{khan_danikhan632auto-gpt-alpacatrader-plugin_2024}, or otherwise transferring funds. Such service providers could require IDs for transactions involving significant sums. IDs could help to reduce the risk of a variety of negative outcomes, by enabling service providers to filter out potentially untrustworthy AI systems (e.g., those that may likely malfunction and make mistaken transactions) and investigate potential incidents. 


\textbf{Contacting humans}: Potential services for contacting humans include crowdwork platforms, telephone services, and online messaging. The ability to contact humans and potentially convince them to take actions in the world could greatly magnify the impact of an AI system. As we discussed in \Cref{sec:scam-calls}, IDs could help to disincentivize and catch misuse of these services. 

\section{Potential ID Implementation}\label{sec:centralized}
We assess a basic ID system that deployers of AI systems could implement. Our discussion in this section will be agnostic to the applications where IDs are used (see \Cref{sec:when-require} for discussion of where ID use may be warranted). 
We also omit discussion of verifiability, given our presentation in \Cref{sec:verifiability} and \Cref{fig:verifiability} of how existing technologies could  enable verifiability. 

We assume a straightforward choice for the identifier, \totalid, where \systemid is the same for all instances of a system\footnote{If the underlying model changes, the system ID should also change. E.g., ChatGPT with a GPT-4 backend is different from ChatGPT with a GPT-3.5 backend.} and \instanceid is unique to a particular instance. It could be useful and straightforward to append other details to the identifier, such as the exact time at which the instance was created. However, we proceed with this choice for simplicity.

\subsection{Assigning IDs}\label{sec:deployer-new-ids}
Although IDs are instance-specific, it may not be feasible for deployers to detect the formation of all new instances. For example, a user could use deployer-run instances $A$ and $B$ as backends for a user-run system $C$, which integrates the outputs of $A$ and $B$ and would evade detection by deployers. We outline some scenarios where deployers could detect new instances and analyze considerations for assigning new IDs.

\textbf{New instances}: Deployers can easily generate new identifiers whenever users create new instances (e.g., a ``new chat'' button).

\textbf{Reloaded instances}: Suppose that a deployer saves instances' states (i.e., the history of interaction) and provides separate functionality for users to reload them. We tentatively suggest that deployers retain the ID of the original instance for the reloaded instance, since some attributes (e.g., prior incidents, context) of the original instance could be relevant for understanding the behaviour of the reloaded instance. Alternatively, if a deployer does decide to assign a new identifier to the reloaded instance, a link to the ID of the original instance would maintain access to the relevant attributes.

\textbf{Output regeneration}: Some deployers provide the ability for a user to regenerate an output in response to previous input. Although not yet implemented, deployers could additionally allow users to create multiple branches (see \Cref{sec:detailed-definition-instances} for further discussion of branches) of interactions, based off of regenerated outputs. More concretely, a user could interact with their instance up until the output $a_t$ at time $t$, regenerate to obtain an output $b_t$, and continue interactions with the respective branches to obtain outputs $(a_{t + 1}, b_{t +1})$, $(a_{t + 2}, b_{t +2})$, etc. Since the two branches have no necessary causal impact on each other (unless they interact with each other or modify a shared object in the world), they should have two separate IDs, with links to the ancestor ID. 

\textbf{Composite systems}: Software frameworks could give users the ability to create new instances (a \textbf{composite}) out of a collection of existing instances \citep{wu_autogen_2023}. If deployers run such software frameworks for users, they could detect the formation of composites. Since the constituent instances can continue running alongside the composite, composites should have new IDs, but could have their IDs linked to their constituents'. 


\textbf{Fine-tuned systems}: Some deployers provide fine-tuning functionality to users. 
Since fine-tuning can change the behaviour of a system, a fine-tuned system $A'$ should have different \systemid than the original system $A$, and link to $A$'s \systemid.
Although it may be useful to include some documentation about the fine-tuning data in an ID, user privacy expectations could complicate such inclusion. Some deployers monitor fine-tuning data only for abuse detection and legal compliance \citep{chrishmsft_data_2023,openai_enterprise_2024}; inclusion of fine-tuning data in an ID likely requires stronger justification.

\subsection{Attributes}\label{sec:deployer-attributes}
We tentatively suggest the following attributes for a deployer to include (or link to) in an ID:
\begin{itemize}
    \item A system card
    \item A database of incidents associated with the system
    \item The IDs of ancestor and descendant instances
\end{itemize}
If instances of future systems persist longer in the world, it may be appropriate to link to a (sub)database of incidents associated with the instance. Implementing links to ancestor and descendant instances, when feasible in the situations discussed in \Cref{sec:deployer-new-ids}, could be a simple way to test the utility of instance-level identification. 

There are likely ways for users to avoid ancestor or descendant links. 
For example, a user could manually copy the inputs of a previous instance to a new instance, avoiding a deployer-provided reloading function. 
More reliable identification of ancestors and descendants could involve more invasive measures, which may only be appropriate in high-stakes domains. For example, a deployer could attempt to analyze the inputs of different instances from the same user for similarities. 

\subsection{Access}\label{sec:deployer–accessibility}
\textbf{Users}: The ID could be accessible through a user interface and sent along with any API requests from a user. If a deployer serves an API to a customer, who subsequently serves an API based on the deployer's system to another user, the deployer would likely have to work with the customer to ensure that the end user receives an ID (see \Cref{sec:maintaining-accessibility} for further discussion).

To help to ensure that secondary parties maintain access to the ID, the deployer could include IDs in watermarks and metadata of media outputs (i.e., text, images, video). Yet, it is unclear how robust such methods may be to user removal \citep{zhang_watermarks_2023,openai_understanding_2024}.

\textbf{Services:} When the AI queries a service, the ID should be sent along with the request. Since requests are text-based (e.g., JSON), including the identifier and (links to) the attributes would be straightforward.

\section{Limitations}\label{sec:limitations}

We discuss limitations of IDs and of our analysis.

\subsection{Misuse of IDs}
If IDs were prevalent but verifiability was not assured, misuse of IDs could be a significant problem. For example, an attacker's agent could masquerade as somebody else's agent to perform illicit actions, such as sending spam or carrying out cyberattacks. Overdependence on such IDs could lead to false accusations against innocent users, who may not have the resources or expertise to defend themselves and point out flaws in the ID system. Ideally, the compromise of an ID should be made clear to all that can access it. IDs should also likely not be the sole or conclusive source of evidence for incident investigations. Concerns about the failure of ID verifiability mirror similar concerns about the reliability of watermarks \citep{gleichauf_digital_2024}. 

\subsection{Decentralized Operation of Agents}
We have not discussed how IDs could be implemented for AI systems which users run themselves, without a deployer. If service providers decide to require IDs, users will need straightforward ID implementation methods. Otherwise, they could be locked out of services or be forced to use a deployer. A potential inspiration is Let's Encrypt \citep{lets_encrypt_lets_2024}, which automated the process of HTTPS certificate creation for websites. A similar service could work with software for running AI systems (e.g., transformers \citep{wolf_transformers_2020}, AutoGen \citep{wu_autogen_2023}) to allow users to request and attach IDs. We leave practical details for future work.

\subsection{Failure to Inform}
Parties could ignore IDs, just as internet users sometimes ignore indicators of untrustworthiness \citep{norris_personality_2021}. Software---or future AI assistants \citep{gabriel_ethics_2024}---could make informed decisions for users in some circumstances, just as browsers reject websites with invalid HTTPS certificates. However, it remains unclear what information would justify rejecting an interaction with an AI system. Other parties could misunderstand or misinterpret an ID's attributes. For example, it can difficult to assess the external validity of evaluation results \citep{fourrier_whats_2023,weidinger_sociotechnical_2023}. 

ID attributes could also be inaccurate. For example, deployers could unintentionally link to an incorrect version of system documentation, or simply fail to update it. OpenAI has occasionally released new versions of GPT-4,\footnote{See \url{https://platform.openai.com/docs/models/gpt-4-turbo-and-gpt-4}.} but has not updated the GPT-4 system card \citep{openai_gpt-4_2023} since the initial release, as of April 2024. As a potential mitigation, regular audits \citep{raji_closing_2020,sharkey_causal_2023} could verify the accuracy of attributes.

\subsection{Bypassing IDs}
As with other types of real-world identity systems, IDs for AI systems cannot wholly prevent undesirable behavior. 
IDs could be lent, or the benefits obtained from ID use---such as purchase of a good---could be transferred. 
Users may hesitate to lend IDs since the borrower could inflict lasting reputational damage on the ID. Yet, similar to the purchase of alcohol for minors, it seems difficult to track or prevent transfer of goods. Future work could assess how likely and how damaging such transfer could be.

\subsection{Privacy Risks}\label{sec:privacy-security}
Although an identifier by itself would not reveal anything about its user, the user's privacy could still be compromised in a number of ways. First, information about the instance could reveal user details. For example, timestamps and details of an instance's activities---as might be contained in incident reports---could help a third-party to identify users. The existence of IDs may engender pressure to attach additional information, which could facilitate further user identification. For example, service abuse could potentially be correlated with certain activity patterns, which could motivate corresponding data collection. Second, if the deployer maintains an internal database linking a system's ID to the corresponding user account, other parties could obtain and misuse the information. Security vulnerabilities could allow attackers to obtain the database. Overreaching governments could force deployers to reveal the users of particular systems.

\subsection{Broader Societal Consequences}\label{sec:broader–societal–consequences}
IDs with additional user information could enable potentially harmful ranking of users. Actors implementing IDs may wish to include such information if it is correlated with undesirable outcomes, such as fraud. Yet, the existence of confounders could lead to instances of some users being unfairly denied access to services, similar to how toxicity classifiers can biased against African American English \citep{sap_risk_2019}. Even if the relationship between information and undesirable outcome was causal, taking such information into account could still be unjustly discriminatory. 

IDs could also enable influence over particular instances and their users. Even if an ID does not contain information about an instance's interactions, external actors could record particular IDs have been involved in interactions, just as cookies can track user activity across websites. Instances that are active for long periods of time could reveal much information about a user's preferences. Actors interacting with instances could try to influence their behaviour through means such as prompting. Advertisers and businesses could try to get instances to purchase or recommend their products or services for their corresponding users. Political actors could try to get instances to influence the voting behaviour of their users. Governments could attempt to enforce rules upon particular instances. Such influence could be strong, especially if personal AI assistants become much more widespread and central to daily interactions \citep{gabriel_ethics_2024}. The specific impacts of this influence deserve further study.

Finally, IDs could create a separate digital channel for interactions between trusted AI systems. 
The vast majority of digital interactions could in the future be between AI systems, especially if future AI assistants handle most of a user's activities \citep{gabriel_ethics_2024}. 
If IDs enable reliable interactions, users may prefer their AI systems to interact only with AI systems that present IDs. Much digital activity could take place between networks of ID-bearing AI instances that trust (and can verify) each other. Entities that trust each other may be able to engage in more productive interactions. As a corollary, instances without IDs may become severely disadvantaged, relative to instances within ID-bearing networks. To avoid such disadvantage, users may have no choice but to submit to the potential negative consequences of IDs we have discussed heretofore in this section. On balance, the consequences of a separate digital channel for AI systems remain unclear.

\section{Related Work}

A growing line of work is concerned with building digital infrastructure to structure how AI systems, especially agents \citep{chan_harms_2023,shavit_practices_2023,gabriel_ethics_2024}, behave and interact. \citet{patil_goex_2024} build a runtime for LLM agents to enable human validation and reversal of actions. \citet{marro_protocol_2024} sketches a protocol for communication between LLMs; the protocol relies on documents to specify structured rules, with natural language as a fallback. \citep{sun_cooperative_2023} argue that decentralized commitment devices will be necessary to allow agents to coordinate with each other. Our work builds upon the preliminary discussion of IDs in \citet{chan_visibility_2024,shavit_practices_2023}. 

Our treatment of verifiability takes inspiration from several Internet and security protocols. HTTPS \citep{fielding_rfc_2022} uses digital certificates \citep{cooper_internet_2008} to allow users to verify the identity of an accessed website. If the certificate is valid, a user's browser uses the information contained in the certificate to establish a secure connection \citep{rescorla_transport_2018}. Websites must obtain such certificates from CAs \citep{cooper_internet_2008}, who verify the identity of the website owner and issue the certificate upon successful identification. Identity verification depends upon public key cryptography \citep{barnes_automatic_2019}, which allows a party to prove ownership of a given, public identifier, otherwise known as a public key. Let's Encrypt is a non-profit CA which automates the process of issuing certificates \citep{lets_encrypt_lets_2024}. HTTPS everywhere was a browser plugin that forced the usage of the HTTPS version of websites whenever it was available; this functionality is now default in modern web browsers \citep{hancock_https_2021}. To design further digital infrastructure for AI systems, other internet and security protocols could be a fruitful source of inspiration \citep{sporny_decentralized_2022,sporny_verifiable_2024,microsoft_introduction_2023}. 

There is much work on verifying the provenance of digital entities. Securing software supply chains \citep{ziv_open_2024,springett_software_2020,souppaya_secure_2022} has become an increasing priority for organizations and governments \citep{the_white_house_national_2023}. A related subject is data provenance, which focuses on tracking the origin and modification of pieces of data in varied contexts, such as production warehouses, scientific research, and environmental protection \citep{buneman_why_2001,simmhan_survey_2005}. Labeling and verifying the provenance of AI-generated outputs has recently received much attention \citep{c2pa_c2pa_2023,zhang_watermarks_2023,gleichauf_digital_2024}.

\section{Conclusion}
To inform crucial decisions about AI interactions, we proposed a framework for creating IDs for AI systems. 
IDs can vary in their attributes, how accessible they are to various parties, and to what extent they are verifiable. IDs could be useful in several scenarios, including for shutting down malfunctioning systems, verifying certifications, and incident investigation. There could be significant demand for IDs from governments, service providers, and users, particularly when systems engage in high-stakes tasks. These actors also have means to incentivize ID usage.
ID implementation seems feasible for deployers, but implementation in decentralized settings will require further study. 

More research is required to understand and address the potential risks of IDs. 
First, more work is required to understand what information is appropriate to attach to IDs, and in what contexts. 
Second, we need to understand better how IDs should be used. 
Third, when ID verifiability fails, there should be ways to limit the impacts of such failure. 
Fourth, the broader societal consequences of introducing IDs require further study.

Limited experimentation with IDs seems justified given the balance of potential benefits and risks. IDs seem most warranted in settings where AI systems could have a large impact upon the world, such as in making financial transactions or contacting real humans. in these cases, service providers could experiment with incentives for ID use. 
For instance, they could rate limit instances without IDs, while still allowing such instances to access services. Deployers could experiment with ID implementation.

IDs have facilitated essential functions across a variety of domains. Incident investigation, allocation of liability, and establishment of trust would be more difficult without IDs for products, organizations, and software systems. As AI systems become increasingly prevalent, deficiencies in such functions could make it more difficult to manage risks from AI systems. If implemented well, IDs for AI systems could enable mechanisms to navigate this emerging world. 

\bibliographystyle{tmlr}
\bibliography{PREPRINT}

\begin{thebibliography}{93}
\providecommand{\natexlab}[1]{#1}
\providecommand{\url}[1]{\texttt{#1}}
\expandafter\ifx\csname urlstyle\endcsname\relax
  \providecommand{\doi}[1]{doi: #1}\else
  \providecommand{\doi}{doi: \begingroup \urlstyle{rm}\Url}\fi

\bibitem[noa(2022)]{noauthor_12_2022}
12 {CFR} § 1026.25 - {Record} retention.
\newblock \emph{Federal Register}, January 2022.

\bibitem[Adept(2022)]{adept_act-1_2022}
Adept.
\newblock {ACT}-1: {Transformer} for {Actions}, September 2022.
\newblock URL \url{https://www.adept.ai/act}.

\bibitem[Agarwal et~al.(2024)Agarwal, Singh, Zhang, Bohnet, Chan, Anand, Abbas, Nova, Co-Reyes, Chu, Behbahani, Faust, and Larochelle]{agarwal_many-shot_2024}
Rishabh Agarwal, Avi Singh, Lei~M. Zhang, Bernd Bohnet, Stephanie Chan, Ankesh Anand, Zaheer Abbas, Azade Nova, John~D. Co-Reyes, Eric Chu, Feryal Behbahani, Aleksandra Faust, and Hugo Larochelle.
\newblock Many-{Shot} {In}-{Context} {Learning}, April 2024.
\newblock URL \url{http://arxiv.org/abs/2404.11018}.
\newblock arXiv:2404.11018 [cs].

\bibitem[Anthropic(2024{\natexlab{a}})]{anthropic_introducing_2024}
Anthropic.
\newblock Introducing the next generation of {Claude}, March 2024{\natexlab{a}}.
\newblock URL \url{https://www.anthropic.com/news/claude-3-family}.

\bibitem[Anthropic(2024{\natexlab{b}})]{anthropic_tool_2024}
Anthropic.
\newblock Tool use (function calling), 2024{\natexlab{b}}.
\newblock URL \url{https://docs.anthropic.com/claude/docs/tool-use}.

\bibitem[Bai et~al.(2022)Bai, Kadavath, Kundu, Askell, Kernion, Jones, Chen, Goldie, Mirhoseini, McKinnon, Chen, Olsson, Olah, Hernandez, Drain, Ganguli, Li, Tran-Johnson, Perez, Kerr, Mueller, Ladish, Landau, Ndousse, Lukosuite, Lovitt, Sellitto, Elhage, Schiefer, Mercado, DasSarma, Lasenby, Larson, Ringer, Johnston, Kravec, Showk, Fort, Lanham, Telleen-Lawton, Conerly, Henighan, Hume, Bowman, Hatfield-Dodds, Mann, Amodei, Joseph, McCandlish, Brown, and Kaplan]{bai_constitutional_2022}
Yuntao Bai, Saurav Kadavath, Sandipan Kundu, Amanda Askell, Jackson Kernion, Andy Jones, Anna Chen, Anna Goldie, Azalia Mirhoseini, Cameron McKinnon, Carol Chen, Catherine Olsson, Christopher Olah, Danny Hernandez, Dawn Drain, Deep Ganguli, Dustin Li, Eli Tran-Johnson, Ethan Perez, Jamie Kerr, Jared Mueller, Jeffrey Ladish, Joshua Landau, Kamal Ndousse, Kamile Lukosuite, Liane Lovitt, Michael Sellitto, Nelson Elhage, Nicholas Schiefer, Noemi Mercado, Nova DasSarma, Robert Lasenby, Robin Larson, Sam Ringer, Scott Johnston, Shauna Kravec, Sheer~El Showk, Stanislav Fort, Tamera Lanham, Timothy Telleen-Lawton, Tom Conerly, Tom Henighan, Tristan Hume, Samuel~R. Bowman, Zac Hatfield-Dodds, Ben Mann, Dario Amodei, Nicholas Joseph, Sam McCandlish, Tom Brown, and Jared Kaplan.
\newblock Constitutional {AI}: {Harmlessness} from {AI} {Feedback}, December 2022.
\newblock URL \url{http://arxiv.org/abs/2212.08073}.
\newblock arXiv:2212.08073 [cs].

\bibitem[Barnes et~al.(2019)Barnes, Hoffman-Andrews, McCarney, and Kasten]{barnes_automatic_2019}
Richard Barnes, Jacob Hoffman-Andrews, Daniel McCarney, and James Kasten.
\newblock Automatic certificate management environment (acme).
\newblock Technical report, 2019.

\bibitem[BBC(2021)]{bbc_post_2021}
BBC.
\newblock Post {Office} scandal explained: {What} the {Horizon} saga is all about.
\newblock April 2021.
\newblock URL \url{https://www.bbc.com/news/business-56718036}.

\bibitem[Beurer-Kellner et~al.(2024)Beurer-Kellner, Balunovic, and Fischer]{beurer-kellner_invariantlabs-aiinvariant_2024}
Luca Beurer-Kellner, Mislav Balunovic, and Marc Fischer.
\newblock invariantlabs-ai/invariant, July 2024.
\newblock URL \url{https://github.com/invariantlabs-ai/invariant}.
\newblock original-date: 2024-05-08T08:57:47Z.

\bibitem[Bickert(2024)]{bickert_our_2024}
Monika Bickert.
\newblock Our {Approach} to {Labeling} {AI}-{Generated} {Content} and {Manipulated} {Media}, April 2024.
\newblock URL \url{https://about.fb.com/news/2024/04/metas-approach-to-labeling-ai-generated-content-and-manipulated-media/}.

\bibitem[Bommasani et~al.(2023)Bommasani, Soylu, Liao, Creel, and Liang]{bommasani_ecosystem_2023}
Rishi Bommasani, Dilara Soylu, Thomas~I. Liao, Kathleen~A. Creel, and Percy Liang.
\newblock Ecosystem {Graphs}: {The} {Social} {Footprint} of {Foundation} {Models}, March 2023.
\newblock URL \url{http://arxiv.org/abs/2303.15772}.
\newblock arXiv:2303.15772 [cs].

\bibitem[Borge et~al.(2017)Borge, Kokoris-Kogias, Jovanovic, Gasser, Gailly, and Ford]{borge_proof--personhood_2017}
Maria Borge, Eleftherios Kokoris-Kogias, Philipp Jovanovic, Linus Gasser, Nicolas Gailly, and Bryan Ford.
\newblock Proof-of-{Personhood}: {Redemocratizing} {Permissionless} {Cryptocurrencies}.
\newblock In \emph{2017 {IEEE} {European} {Symposium} on {Security} and {Privacy} {Workshops} ({EuroS}\&{PW})}, pp.\  23--26, April 2017.
\newblock \doi{10.1109/EuroSPW.2017.46}.
\newblock URL \url{https://ieeexplore.ieee.org/abstract/document/7966966}.

\bibitem[Buiten et~al.(2023)Buiten, de~Streel, and Peitz]{buiten_law_2023}
Miriam Buiten, Alexandre de~Streel, and Martin Peitz.
\newblock The law and economics of {AI} liability.
\newblock \emph{Computer Law \& Security Review}, 48:\penalty0 105794, April 2023.
\newblock ISSN 0267-3649.
\newblock \doi{10.1016/j.clsr.2023.105794}.
\newblock URL \url{https://www.sciencedirect.com/science/article/pii/S0267364923000055}.

\bibitem[Buiten(2024)]{buiten_product_2024}
Miriam~C. Buiten.
\newblock Product liability for defective {AI}.
\newblock \emph{European Journal of Law and Economics}, February 2024.
\newblock ISSN 1572-9990.
\newblock \doi{10.1007/s10657-024-09794-z}.
\newblock URL \url{https://doi.org/10.1007/s10657-024-09794-z}.

\bibitem[Buneman et~al.(2001)Buneman, Khanna, and Wang-Chiew]{buneman_why_2001}
Peter Buneman, Sanjeev Khanna, and Tan Wang-Chiew.
\newblock Why and {Where}: {A} {Characterization} of {Data} {Provenance}.
\newblock October 2001.
\newblock \doi{10.1007/3-540-44503-X_20}.
\newblock URL \url{https://link.springer.com/chapter/10.1007/3-540-44503-X_20}.

\bibitem[C2PA(2023)]{c2pa_c2pa_2023}
C2PA.
\newblock {C2PA} {Technical} {Specification}, 2023.
\newblock URL \url{https://c2pa.org/specifications/specifications/1.3/specs/C2PA_Specification.html#_introduction}.

\bibitem[Chan et~al.(2023)Chan, Salganik, Markelius, Pang, Rajkumar, Krasheninnikov, Langosco, He, Duan, Carroll, Lin, Mayhew, Collins, Molamohammadi, Burden, Zhao, Rismani, Voudouris, Bhatt, Weller, Krueger, and Maharaj]{chan_harms_2023}
Alan Chan, Rebecca Salganik, Alva Markelius, Chris Pang, Nitarshan Rajkumar, Dmitrii Krasheninnikov, Lauro Langosco, Zhonghao He, Yawen Duan, Micah Carroll, Michelle Lin, Alex Mayhew, Katherine Collins, Maryam Molamohammadi, John Burden, Wanru Zhao, Shalaleh Rismani, Konstantinos Voudouris, Umang Bhatt, Adrian Weller, David Krueger, and Tegan Maharaj.
\newblock Harms from {Increasingly} {Agentic} {Algorithmic} {Systems}.
\newblock In \emph{Proceedings of the 2023 {ACM} {Conference} on {Fairness}, {Accountability}, and {Transparency}}, {FAccT} '23, pp.\  651--666, New York, NY, USA, June 2023. Association for Computing Machinery.
\newblock ISBN 9798400701924.
\newblock \doi{10.1145/3593013.3594033}.
\newblock URL \url{https://dl.acm.org/doi/10.1145/3593013.3594033}.

\bibitem[Chan et~al.(2024)Chan, Ezell, Kaufmann, Wei, Hammond, Bradley, Bluemke, Rajkumar, Krueger, Kolt, Heim, and Anderljung]{chan_visibility_2024}
Alan Chan, Carson Ezell, Max Kaufmann, Kevin Wei, Lewis Hammond, Herbie Bradley, Emma Bluemke, Nitarshan Rajkumar, David Krueger, Noam Kolt, Lennart Heim, and Markus Anderljung.
\newblock Visibility into {AI} {Agents}, February 2024.
\newblock URL \url{http://arxiv.org/abs/2401.13138}.
\newblock arXiv:2401.13138 [cs].

\bibitem[ChrisHMSFT et~al.(2023)ChrisHMSFT, PatrickFarley, mrbullwinkle, eric urban, and aahill]{chrishmsft_data_2023}
ChrisHMSFT, PatrickFarley, mrbullwinkle, eric urban, and aahill.
\newblock Data, privacy, and security for {Azure} {OpenAI} {Service} - {Azure} {AI} services, June 2023.
\newblock URL \url{https://learn.microsoft.com/en-us/legal/cognitive-services/openai/data-privacy}.

\bibitem[Cohen et~al.(2024)Cohen, Bitton, and Nassi]{cohen_here_2024}
Stav Cohen, Ron Bitton, and Ben Nassi.
\newblock Here {Comes} {The} {AI} {Worm}: {Unleashing} {Zero}-click {Worms} that {Target} {GenAI}-{Powered} {Applications}, March 2024.
\newblock URL \url{http://arxiv.org/abs/2403.02817}.
\newblock arXiv:2403.02817 [cs].

\bibitem[Cooper et~al.(2008)Cooper, Santesson, Farrell, Boeyen, Housley, and Polk]{cooper_internet_2008}
David Cooper, Stefan Santesson, Stephen Farrell, Sharon Boeyen, Russell Housley, and William Polk.
\newblock Internet {X}.509 public key infrastructure certificate and certificate revocation list ({CRL}) profile.
\newblock Technical report, 2008.

\bibitem[De~La~Garza(2020)]{de_la_garza_states_2020}
Alejandro De~La~Garza.
\newblock States' {Automated} {Systems} {Are} {Trapping} {Citizens} in {Bureaucratic} {Nightmares} {With} {Their} {Lives} on the {Line}.
\newblock \emph{TIME}, May 2020.
\newblock URL \url{https://time.com/5840609/algorithm-unemployment/}.

\bibitem[Egan \& Heim(2023)Egan and Heim]{egan_oversight_2023}
Janet Egan and Lennart Heim.
\newblock Oversight for {Frontier} {AI} through a {Know}-{Your}-{Customer} {Scheme} for {Compute} {Providers}, October 2023.
\newblock URL \url{http://arxiv.org/abs/2310.13625}.
\newblock arXiv:2310.13625 [cs].

\bibitem[Encrypt(2024)]{lets_encrypt_lets_2024}
Let's Encrypt.
\newblock Let's {Encrypt} {Stats} - {Let}'s {Encrypt}, January 2024.
\newblock URL \url{https://letsencrypt.org/stats/}.

\bibitem[{Epoch}(2023)]{epoch_key_2023}
{Epoch}.
\newblock Key trends and figures in {Machine} {Learning}, 2023.
\newblock URL \url{https://epochai.org/trends}.

\bibitem[Erdil \& Besiroglu(2023)Erdil and Besiroglu]{erdil_algorithmic_2023}
Ege Erdil and Tamay Besiroglu.
\newblock Algorithmic progress in computer vision, 2023.
\newblock \_eprint: 2212.05153.

\bibitem[Fielding et~al.(2022)Fielding, Nottingham, and Reschke]{fielding_rfc_2022}
R~Fielding, M~Nottingham, and J~Reschke.
\newblock {RFC} 9110: {HTTP} semantics, 2022.

\bibitem[Fourrier et~al.(2023)Fourrier, Habib, Launay, and Wolf]{fourrier_whats_2023}
Clémentine Fourrier, Nathan Habib, Julien Launay, and Thomas Wolf.
\newblock What's going on with the {Open} {LLM} {Leaderboard}?, June 2023.
\newblock URL \url{https://huggingface.co/blog/open-llm-leaderboard-mmlu}.

\bibitem[Furuta et~al.(2024)Furuta, Matsuo, Faust, and Gur]{furuta_exposing_2024}
Hiroki Furuta, Yutaka Matsuo, Aleksandra Faust, and Izzeddin Gur.
\newblock Exposing {Limitations} of {Language} {Model} {Agents} in {Sequential}-{Task} {Compositions} on the {Web}, February 2024.
\newblock URL \url{http://arxiv.org/abs/2311.18751}.
\newblock arXiv:2311.18751 [cs].

\bibitem[Gabriel et~al.(2024)Gabriel, Manzini, Keeling, Hendricks, Rieser, Iqbal, Tomašev, Ktena, Kenton, Rodriguez, El-Sayed, Brown, Akbulut, Trask, Hughes, Bergman, Shelby, Marchal, Griffin, Mateos-Garcia, Weidinger, Street, Lange, Ingerman, Lentz, Enger, Barakat, Krakovna, Siy, Kurth-Nelson, McCroskery, Bolina, Law, Shanahan, Alberts, Balle, de~Haas, Ibitoye, Dafoe, Goldberg, Krier, Reese, Witherspoon, Hawkins, Rauh, Wallace, Franklin, Goldstein, Lehman, Klenk, Vallor, Biles, Morris, King, Arcas, Isaac, and Manyika]{gabriel_ethics_2024}
Iason Gabriel, Arianna Manzini, Geoff Keeling, Lisa~Anne Hendricks, Verena Rieser, Hasan Iqbal, Nenad Tomašev, Ira Ktena, Zachary Kenton, Mikel Rodriguez, Seliem El-Sayed, Sasha Brown, Canfer Akbulut, Andrew Trask, Edward Hughes, A.~Stevie Bergman, Renee Shelby, Nahema Marchal, Conor Griffin, Juan Mateos-Garcia, Laura Weidinger, Winnie Street, Benjamin Lange, Alex Ingerman, Alison Lentz, Reed Enger, Andrew Barakat, Victoria Krakovna, John~Oliver Siy, Zeb Kurth-Nelson, Amanda McCroskery, Vijay Bolina, Harry Law, Murray Shanahan, Lize Alberts, Borja Balle, Sarah de~Haas, Yetunde Ibitoye, Allan Dafoe, Beth Goldberg, Sébastien Krier, Alexander Reese, Sims Witherspoon, Will Hawkins, Maribeth Rauh, Don Wallace, Matija Franklin, Josh~A. Goldstein, Joel Lehman, Michael Klenk, Shannon Vallor, Courtney Biles, Meredith~Ringel Morris, Helen King, Blaise Agüera~y Arcas, William Isaac, and James Manyika.
\newblock The {Ethics} of {Advanced} {AI} {Assistants}, April 2024.
\newblock URL \url{http://arxiv.org/abs/2404.16244}.
\newblock arXiv:2404.16244 [cs].

\bibitem[Gebru et~al.(2021)Gebru, Morgenstern, Vecchione, Vaughan, Wallach, Iii, and Crawford]{gebru_datasheets_2021}
Timnit Gebru, Jamie Morgenstern, Briana Vecchione, Jennifer~Wortman Vaughan, Hanna Wallach, Hal~Daumé Iii, and Kate Crawford.
\newblock Datasheets for datasets.
\newblock \emph{Communications of the ACM}, 64\penalty0 (12):\penalty0 86--92, December 2021.
\newblock ISSN 0001-0782, 1557-7317.
\newblock \doi{10.1145/3458723}.
\newblock URL \url{https://dl.acm.org/doi/10.1145/3458723}.

\bibitem[Gilbert et~al.(2023)Gilbert, Lambert, Dean, Zick, Snoswell, and Mehta]{gilbert_reward_2023}
Thomas~Krendl Gilbert, Nathan Lambert, Sarah Dean, Tom Zick, Aaron Snoswell, and Soham Mehta.
\newblock Reward {Reports} for {Reinforcement} {Learning}.
\newblock In \emph{Proceedings of the 2023 {AAAI}/{ACM} {Conference} on {AI}, {Ethics}, and {Society}}, {AIES} '23, pp.\  84--130, New York, NY, USA, August 2023. Association for Computing Machinery.
\newblock ISBN 9798400702310.
\newblock \doi{10.1145/3600211.3604698}.
\newblock URL \url{https://dl.acm.org/doi/10.1145/3600211.3604698}.

\bibitem[Gleichauf \& Geer(2024)Gleichauf and Geer]{gleichauf_digital_2024}
Bob Gleichauf and Dan Geer.
\newblock Digital {Watermarks} {Are} {Not} {Ready} for {Large} {Language} {Models}, February 2024.
\newblock URL \url{https://www.lawfaremedia.org/article/digital-watermarks-are-not-ready-for-large-language-models}.

\bibitem[Greshake et~al.(2023)Greshake, Abdelnabi, Mishra, Endres, Holz, and Fritz]{greshake_not_2023}
Kai Greshake, Sahar Abdelnabi, Shailesh Mishra, Christoph Endres, Thorsten Holz, and Mario Fritz.
\newblock Not what you've signed up for: {Compromising} {Real}-{World} {LLM}-{Integrated} {Applications} with {Indirect} {Prompt} {Injection}, May 2023.
\newblock URL \url{http://arxiv.org/abs/2302.12173}.
\newblock arXiv:2302.12173 [cs].

\bibitem[Gur et~al.(2023)Gur, Nachum, Miao, Safdari, Huang, Chowdhery, Narang, Fiedel, and Faust]{gur_understanding_2023}
Izzeddin Gur, Ofir Nachum, Yingjie Miao, Mustafa Safdari, Austin Huang, Aakanksha Chowdhery, Sharan Narang, Noah Fiedel, and Aleksandra Faust.
\newblock Understanding {HTML} with {Large} {Language} {Models}, May 2023.
\newblock URL \url{http://arxiv.org/abs/2210.03945}.
\newblock arXiv:2210.03945 [cs].

\bibitem[Gur et~al.(2024)Gur, Furuta, Huang, Safdari, Matsuo, Eck, and Faust]{gur_real-world_2024}
Izzeddin Gur, Hiroki Furuta, Austin Huang, Mustafa Safdari, Yutaka Matsuo, Douglas Eck, and Aleksandra Faust.
\newblock A {Real}-{World} {WebAgent} with {Planning}, {Long} {Context} {Understanding}, and {Program} {Synthesis}, February 2024.
\newblock URL \url{http://arxiv.org/abs/2307.12856}.
\newblock arXiv:2307.12856 [cs].

\bibitem[Hancock(2021)]{hancock_https_2021}
Alexis Hancock.
\newblock {HTTPS} {Is} {Actually} {Everywhere}, September 2021.
\newblock URL \url{https://www.eff.org/deeplinks/2021/09/https-actually-everywhere}.

\bibitem[Ho et~al.(2024)Ho, Besiroglu, Erdil, Owen, Rahman, Guo, Atkinson, Thompson, and Sevilla]{ho_algorithmic_2024}
Anson Ho, Tamay Besiroglu, Ege Erdil, David Owen, Robi Rahman, Zifan~Carl Guo, David Atkinson, Neil Thompson, and Jaime Sevilla.
\newblock Algorithmic progress in language models, March 2024.
\newblock URL \url{https://arxiv.org/abs/2403.05812v1}.

\bibitem[Hoffmann et~al.(2022)Hoffmann, Borgeaud, Mensch, Buchatskaya, Cai, Rutherford, Casas, Hendricks, Welbl, Clark, Hennigan, Noland, Millican, Driessche, Damoc, Guy, Osindero, Simonyan, Elsen, Rae, Vinyals, and Sifre]{hoffmann_training_2022}
Jordan Hoffmann, Sebastian Borgeaud, Arthur Mensch, Elena Buchatskaya, Trevor Cai, Eliza Rutherford, Diego de~Las Casas, Lisa~Anne Hendricks, Johannes Welbl, Aidan Clark, Tom Hennigan, Eric Noland, Katie Millican, George van~den Driessche, Bogdan Damoc, Aurelia Guy, Simon Osindero, Karen Simonyan, Erich Elsen, Jack~W. Rae, Oriol Vinyals, and Laurent Sifre.
\newblock Training {Compute}-{Optimal} {Large} {Language} {Models}, March 2022.
\newblock URL \url{http://arxiv.org/abs/2203.15556}.
\newblock arXiv:2203.15556 [cs].

\bibitem[House(2023)]{the_white_house_national_2023}
The~White House.
\newblock National {Cybersecurity} {Strategy}.
\newblock Technical report, March 2023.

\bibitem[Jimenez et~al.(2024)Jimenez, Yang, Wettig, Yao, Pei, Press, and Narasimhan]{jimenez_swe-bench_2024}
Carlos~E. Jimenez, John Yang, Alexander Wettig, Shunyu Yao, Kexin Pei, Ofir Press, and Karthik Narasimhan.
\newblock {SWE}-bench: {Can} {Language} {Models} {Resolve} {Real}-{World} {GitHub} {Issues}?, April 2024.
\newblock URL \url{http://arxiv.org/abs/2310.06770}.
\newblock arXiv:2310.06770 [cs].

\bibitem[Khan et~al.(2024)Khan, Mishra, and Toren-Herrinton]{khan_danikhan632auto-gpt-alpacatrader-plugin_2024}
Daniyal Khan, Gyanendra Mishra, and Ari Toren-Herrinton.
\newblock danikhan632/{Auto}-{GPT}-{AlpacaTrader}-{Plugin}, March 2024.
\newblock URL \url{https://github.com/danikhan632/Auto-GPT-AlpacaTrader-Plugin}.
\newblock original-date: 2023-04-25T16:51:50Z.

\bibitem[Kinniment et~al.(2023)Kinniment, Sato, Du, Goodrich, Hasin, Chan, Miles, Lin, Wijk, Burget, Ho, Barnes, and Christiano]{kinniment_evaluating_2023}
Megan Kinniment, Lucas Jun~Koba Sato, Haoxing Du, Brian Goodrich, Max Hasin, Lawrence Chan, Luke~Harold Miles, Tao~R. Lin, Hjalmar Wijk, Joel Burget, Aaron Ho, Elizabeth Barnes, and Paul Christiano.
\newblock Evaluating {Language}-{Model} {Agents} on {Realistic} {Autonomous} {Tasks}, July 2023.
\newblock URL \url{https://evals.alignment.org/Evaluating_LMAs_Realistic_Tasks.pdf}.

\bibitem[Kolt(2024)]{kolt_governing_2024}
Noam Kolt.
\newblock Governing {AI} {Agents}, April 2024.
\newblock URL \url{https://papers.ssrn.com/abstract=4772956}.

\bibitem[Korenhof et~al.(2014)Korenhof, Koning, Alpár, and Hoepman]{korenhof_abc_2014}
P.E.I. Korenhof, Merel Koning, Gergely Alpár, and J.H. Hoepman.
\newblock The {ABC} of {ABC}: {An} analysis of attribute-based credentials in the light of data protection, privacy and identity.
\newblock \emph{Internet, Law and Politics}, 10:\penalty0 357--374, July 2014.
\newblock Place: Barcelona Publisher: Huygens Editorial.

\bibitem[Liu et~al.(2024)Liu, Pan, Lu, Li, Hu, Wen, King, and Yu]{liu_survey_2024}
Aiwei Liu, Leyi Pan, Yijian Lu, Jingjing Li, Xuming Hu, Lijie Wen, Irwin King, and Philip~S. Yu.
\newblock A {Survey} of {Text} {Watermarking} in the {Era} of {Large} {Language} {Models}, January 2024.
\newblock URL \url{http://arxiv.org/abs/2312.07913}.
\newblock arXiv:2312.07913 [cs].

\bibitem[Liu et~al.(2023)Liu, Yu, Zhang, Xu, Lei, Lai, Gu, Ding, Men, Yang, Zhang, Deng, Zeng, Du, Zhang, Shen, Zhang, Su, Sun, Huang, Dong, and Tang]{liu_agentbench_2023}
Xiao Liu, Hao Yu, Hanchen Zhang, Yifan Xu, Xuanyu Lei, Hanyu Lai, Yu~Gu, Hangliang Ding, Kaiwen Men, Kejuan Yang, Shudan Zhang, Xiang Deng, Aohan Zeng, Zhengxiao Du, Chenhui Zhang, Sheng Shen, Tianjun Zhang, Yu~Su, Huan Sun, Minlie Huang, Yuxiao Dong, and Jie Tang.
\newblock {AgentBench}: {Evaluating} {LLMs} as {Agents}, October 2023.
\newblock URL \url{http://arxiv.org/abs/2308.03688}.
\newblock arXiv:2308.03688 [cs].

\bibitem[Marro(2024)]{marro_protocol_2024}
Samuele Marro.
\newblock A {Protocol} {Sketch} {For} {LLM} {Communication}, April 2024.
\newblock URL \url{https://samuelemarro.it/blog/2024/a-protocol-for-llm/}.

\bibitem[Mialon et~al.(2023)Mialon, Fourrier, Swift, Wolf, LeCun, and Scialom]{mialon_gaia_2023}
Grégoire Mialon, Clémentine Fourrier, Craig Swift, Thomas Wolf, Yann LeCun, and Thomas Scialom.
\newblock {GAIA}: a benchmark for {General} {AI} {Assistants}, November 2023.
\newblock URL \url{https://arxiv.org/abs/2311.12983v1}.

\bibitem[Microsoft(2023)]{microsoft_introduction_2023}
Microsoft.
\newblock Introduction to {Microsoft} {Entra} {Verified} {ID}, November 2023.
\newblock URL \url{https://learn.microsoft.com/en-us/entra/verified-id/decentralized-identifier-overview}.

\bibitem[Mitchell et~al.(2019)Mitchell, Wu, Zaldivar, Barnes, Vasserman, Hutchinson, Spitzer, Raji, and Gebru]{mitchell_model_2019}
Margaret Mitchell, Simone Wu, Andrew Zaldivar, Parker Barnes, Lucy Vasserman, Ben Hutchinson, Elena Spitzer, Inioluwa~Deborah Raji, and Timnit Gebru.
\newblock Model {Cards} for {Model} {Reporting}.
\newblock In \emph{Proceedings of the {Conference} on {Fairness}, {Accountability}, and {Transparency}}, pp.\  220--229, January 2019.
\newblock \doi{10.1145/3287560.3287596}.
\newblock URL \url{http://arxiv.org/abs/1810.03993}.
\newblock arXiv:1810.03993 [cs].

\bibitem[Norris \& Brookes(2021)Norris and Brookes]{norris_personality_2021}
Gareth Norris and Alexandra Brookes.
\newblock Personality, emotion and individual differences in response to online fraud.
\newblock \emph{Personality and Individual Differences}, 169:\penalty0 109847, February 2021.
\newblock ISSN 0191-8869.
\newblock \doi{10.1016/j.paid.2020.109847}.
\newblock URL \url{https://www.sciencedirect.com/science/article/pii/S0191886920300374}.

\bibitem[OpenAI(2023{\natexlab{a}})]{openai_chatgpt_2023}
OpenAI.
\newblock {ChatGPT} plugins, 2023{\natexlab{a}}.
\newblock URL \url{https://openai.com/blog/chatgpt-plugins}.

\bibitem[OpenAI(2023{\natexlab{b}})]{openai_gpt-4_2023}
OpenAI.
\newblock {GPT}-4 {Technical} {Report}, March 2023{\natexlab{b}}.
\newblock URL \url{http://arxiv.org/abs/2303.08774}.
\newblock arXiv:2303.08774 [cs].

\bibitem[OpenAI(2024{\natexlab{a}})]{openai_enterprise_2024}
OpenAI.
\newblock Enterprise privacy, January 2024{\natexlab{a}}.
\newblock URL \url{https://openai.com/enterprise-privacy}.

\bibitem[OpenAI(2024{\natexlab{b}})]{openai_introducing_2024}
OpenAI.
\newblock Introducing the {GPT} {Store}, January 2024{\natexlab{b}}.
\newblock URL \url{https://openai.com/blog/introducing-the-gpt-store}.

\bibitem[OpenAI(2024{\natexlab{c}})]{openai_understanding_2024}
OpenAI.
\newblock Understanding the source of what we see and hear online, May 2024{\natexlab{c}}.
\newblock URL \url{https://openai.com/index/understanding-the-source-of-what-we-see-and-hear-online/}.

\bibitem[Parliament(2024)]{european_parliament_artificial_2024}
European Parliament.
\newblock Artificial {Intelligence} {Act}, March 2024.

\bibitem[Patil et~al.(2024)Patil, Zhang, Fang, C., Huang, Hao, Casado, Gonzalez, Popa, and Stoica]{patil_goex_2024}
Shishir~G. Patil, Tianjun Zhang, Vivian Fang, Noppapon C., Roy Huang, Aaron Hao, Martin Casado, Joseph~E. Gonzalez, Raluca~Ada Popa, and Ion Stoica.
\newblock {GoEX}: {Perspectives} and {Designs} {Towards} a {Runtime} for {Autonomous} {LLM} {Applications}, April 2024.
\newblock URL \url{http://arxiv.org/abs/2404.06921}.
\newblock arXiv:2404.06921 [cs].

\bibitem[Raji et~al.(2020)Raji, Smart, White, Mitchell, Gebru, Hutchinson, Smith-Loud, Theron, and Barnes]{raji_closing_2020}
Inioluwa~Deborah Raji, Andrew Smart, Rebecca~N. White, Margaret Mitchell, Timnit Gebru, Ben Hutchinson, Jamila Smith-Loud, Daniel Theron, and Parker Barnes.
\newblock Closing the {AI} accountability gap: defining an end-to-end framework for internal algorithmic auditing.
\newblock In \emph{Proceedings of the 2020 {Conference} on {Fairness}, {Accountability}, and {Transparency}}, {FAT}* '20, pp.\  33--44, New York, NY, USA, January 2020. Association for Computing Machinery.
\newblock ISBN 978-1-4503-6936-7.
\newblock \doi{10.1145/3351095.3372873}.
\newblock URL \url{https://dl.acm.org/doi/10.1145/3351095.3372873}.

\bibitem[Raji et~al.(2022)Raji, Kumar, Horowitz, and Selbst]{raji_fallacy_2022}
Inioluwa~Deborah Raji, I.~Elizabeth Kumar, Aaron Horowitz, and Andrew Selbst.
\newblock The {Fallacy} of {AI} {Functionality}.
\newblock In \emph{2022 {ACM} {Conference} on {Fairness}, {Accountability}, and {Transparency}}, pp.\  959--972, Seoul Republic of Korea, June 2022. ACM.
\newblock ISBN 978-1-4503-9352-2.
\newblock \doi{10.1145/3531146.3533158}.
\newblock URL \url{https://dl.acm.org/doi/10.1145/3531146.3533158}.

\bibitem[Rescorla(2018)]{rescorla_transport_2018}
Eric Rescorla.
\newblock The transport layer security ({TLS}) protocol version 1.3.
\newblock Technical report, 2018.

\bibitem[Richards(2023)]{richards_auto-gpt_2023}
Toran~Bruce Richards.
\newblock Auto-{GPT}: {An} {Autonomous} {GPT}-4 {Experiment}, April 2023.
\newblock URL \url{https://github.com/Significant-Gravitas/Auto-GPT}.
\newblock original-date: 2023-03-16T09:21:07Z.

\bibitem[Sap et~al.(2019)Sap, Card, Gabriel, Choi, and Smith]{sap_risk_2019}
Maarten Sap, Dallas Card, Saadia Gabriel, Yejin Choi, and Noah~A. Smith.
\newblock The {Risk} of {Racial} {Bias} in {Hate} {Speech} {Detection}.
\newblock In Anna Korhonen, David Traum, and Lluís Màrquez (eds.), \emph{Proceedings of the 57th {Annual} {Meeting} of the {Association} for {Computational} {Linguistics}}, pp.\  1668--1678, Florence, Italy, July 2019. Association for Computational Linguistics.
\newblock \doi{10.18653/v1/P19-1163}.
\newblock URL \url{https://aclanthology.org/P19-1163}.

\bibitem[Shanahan et~al.(2023)Shanahan, McDonell, and Reynolds]{shanahan_role-play_2023}
Murray Shanahan, Kyle McDonell, and Laria Reynolds.
\newblock Role-{Play} with {Large} {Language} {Models}, May 2023.
\newblock URL \url{http://arxiv.org/abs/2305.16367}.
\newblock arXiv:2305.16367 [cs].

\bibitem[Sharkey et~al.(2023)Sharkey, {Clíodhna Ní Ghuidhir}, {Dan Braun}, {Jérémy Scheurer}, {Mikita Balesni}, {Lucius Bushnaq}, {Charlotte Stix}, and {Marius Hobbhahn}]{sharkey_causal_2023}
Lee Sharkey, {Clíodhna Ní Ghuidhir}, {Dan Braun}, {Jérémy Scheurer}, {Mikita Balesni}, {Lucius Bushnaq}, {Charlotte Stix}, and {Marius Hobbhahn}.
\newblock A {Causal} {Framework} for {AI} {Regulation} and {Auditing}, November 2023.
\newblock URL \url{https://static1.squarespace.com/static/6461e2a5c6399341bcfc84a5/t/654bc268049d687cecac24d8/1699463818729/auditing_framework_web.pdf}.

\bibitem[Shavit et~al.(2023)Shavit, Agarwal, Brundage, Adler, O’Keefe, Campbell, Lee, Mishkin, Eloundou, Hickey, Slama, Ahmad, McMillan, Beutel, Passos, and Robinson]{shavit_practices_2023}
Yonadav Shavit, Sandhini Agarwal, Miles Brundage, Steven Adler, Cullen O’Keefe, Rosie Campbell, Teddy Lee, Pamela Mishkin, Tyna Eloundou, Alan Hickey, Katarina Slama, Lama Ahmad, Paul McMillan, Alex Beutel, Alexandre Passos, and David~G. Robinson.
\newblock Practices for {Governing} {Agentic} {AI} {Systems}, 2023.

\bibitem[Shet(2014)]{shet_are_2014}
Vinay Shet.
\newblock Are you a robot? {Introducing} "{No} {CAPTCHA} {reCAPTCHA}", December 2014.
\newblock URL \url{https://developers.google.com/search/blog/2014/12/are-you-robot-introducing-no-captcha}.

\bibitem[Shevlane et~al.(2023)Shevlane, Farquhar, Garfinkel, Phuong, Whittlestone, Leung, Kokotajlo, Marchal, Anderljung, Kolt, Ho, Siddarth, Avin, Hawkins, Kim, Gabriel, Bolina, Clark, Bengio, Christiano, and Dafoe]{shevlane_model_2023}
Toby Shevlane, Sebastian Farquhar, Ben Garfinkel, Mary Phuong, Jess Whittlestone, Jade Leung, Daniel Kokotajlo, Nahema Marchal, Markus Anderljung, Noam Kolt, Lewis Ho, Divya Siddarth, Shahar Avin, Will Hawkins, Been Kim, Iason Gabriel, Vijay Bolina, Jack Clark, Yoshua Bengio, Paul Christiano, and Allan Dafoe.
\newblock Model evaluation for extreme risks, May 2023.
\newblock URL \url{https://arxiv.org/abs/2305.15324v2}.

\bibitem[Significant-Gravitas(2024)]{significant-gravitas_auto-gpt-plugins_2024}
Significant-Gravitas.
\newblock Auto-{GPT}-{Plugins}, 2024.
\newblock URL \url{https://github.com/Significant-Gravitas/Auto-GPT-Plugins}.

\bibitem[Simmhan et~al.(2005)Simmhan, Plale, Gannon, and {others}]{simmhan_survey_2005}
Yogesh~L Simmhan, Beth Plale, Dennis Gannon, and {others}.
\newblock A survey of data provenance techniques.
\newblock \emph{Computer Science Department, Indiana University, Bloomington IN}, 47405:\penalty0 69, 2005.

\bibitem[Souppaya et~al.(2022)Souppaya, Scarfone, and Dodson]{souppaya_secure_2022}
Murugiah Souppaya, Karen Scarfone, and Donna Dodson.
\newblock Secure {Software} {Development} {Framework} ({SSDF}) {Version} 1.1: {Recommendations} for {Mitigating} the {Risk} of {Software} {Vulnerabilities}.
\newblock Technical report, 2022.

\bibitem[Sporny et~al.(2022)Sporny, Longley, Reed, Sabadello, Steele, and Allen]{sporny_decentralized_2022}
Manu Sporny, Dave Longley, Drummond Reed, Markus Sabadello, Orie Steele, and Christopher Allen.
\newblock Decentralized {Identifiers} ({DiDs}) v1.0.
\newblock Technical report, W3C, July 2022.

\bibitem[Sporny et~al.(2024)Sporny, Longley, Chadwick, and Steele]{sporny_verifiable_2024}
Manu Sporny, Dave Longley, David Chadwick, and Orie Steele.
\newblock Verifiable {Credentials} {Data} {Model} v2.0.
\newblock Technical report, W3C, May 2024.

\bibitem[Springett et~al.(2020)Springett, {Dave Russo}, {Garret Fick}, {JC Herz}, {John Scott}, {Mark Symons}, {Pruthvi Nallapareddy}, and {Bryan Garcia}]{springett_software_2020}
Steve Springett, {Dave Russo}, {Garret Fick}, {JC Herz}, {John Scott}, {Mark Symons}, {Pruthvi Nallapareddy}, and {Bryan Garcia}.
\newblock Software {Component} {Verification} {Standard}.
\newblock Technical report, OWASP, June 2020.
\newblock URL \url{https://github.com/OWASP/Software-Component-Verification-Standard}.
\newblock original-date: 2019-08-28T15:27:27Z.

\bibitem[Sun et~al.(2023)Sun, Crapis, Stephenson, Monnot, Thiery, and Passerat-Palmbach]{sun_cooperative_2023}
Xinyuan Sun, Davide Crapis, Matt Stephenson, Barnabé Monnot, Thomas Thiery, and Jonathan Passerat-Palmbach.
\newblock Cooperative {AI} via {Decentralized} {Commitment} {Devices}, November 2023.
\newblock URL \url{http://arxiv.org/abs/2311.07815}.
\newblock arXiv:2311.07815 [cs].

\bibitem[Tao et~al.(2023)Tao, T~V, Shlapentokh-Rothman, and Hoiem]{tao_webwise_2023}
Heyi Tao, Sethuraman T~V, Michal Shlapentokh-Rothman, and Derek Hoiem.
\newblock {WebWISE}: {Web} {Interface} {Control} and {Sequential} {Exploration} with {Large} {Language} {Models}, October 2023.
\newblock URL \url{http://arxiv.org/abs/2310.16042}.
\newblock arXiv:2310.16042 [cs].

\bibitem[Trask et~al.(2023)Trask, Sukumar, Kalliokoski, Farkas, Ezenwaka, Popa, Mitchell, Hrebenach, Muraru, Junior, Bejan, Mishra, Ngong, Bandy, Stahl, Cardonnet, Trask, Nguyen, Dang, Veen, Eng, Strahm, Ayre, Jay, Lytvyn, Kyemenu-Sarsah, Chung, Smith, S, Falcon, Gupta, Gabriel, Milea, Thoraldson, Porto, Cebere, Gorana, and Reza]{trask_how_2023}
Andrew Trask, Akshay Sukumar, Antti Kalliokoski, Bennett Farkas, Callis Ezenwaka, Carmen Popa, Curtis Mitchell, Dylan Hrebenach, George-Cristian Muraru, Ionesio Junior, Irina Bejan, Ishan Mishra, Ivoline Ngong, Jack Bandy, Jess Stahl, Julian Cardonnet, Kellye Trask, Khoa Nguyen, Kien Dang, Koen van~der Veen, Kyoko Eng, Lacey Strahm, Laura Ayre, Madhava Jay, Oleksandr Lytvyn, Osam Kyemenu-Sarsah, Peter Chung, Peter Smith, Rasswanth S, Ronnie Falcon, Shubham Gupta, Stephen Gabriel, Teo Milea, Theresa Thoraldson, Thiago Porto, Tudor Cebere, Yash Gorana, and Zarreen Reza.
\newblock How to audit an {AI} model owned by someone else (part 1), July 2023.
\newblock URL \url{https://blog.openmined.org/ai-audit-part-1/}.

\bibitem[Trask et~al.(2024)Trask, Bluemke, Collins, Drexler, Cuervas-Mons, Gabriel, Dafoe, and Isaac]{trask_beyond_2024}
Andrew Trask, Emma Bluemke, Teddy Collins, Ben Garfinkel~Eric Drexler, Claudia~Ghezzou Cuervas-Mons, Iason Gabriel, Allan Dafoe, and William Isaac.
\newblock Beyond {Privacy} {Trade}-offs with {Structured} {Transparency}, March 2024.
\newblock URL \url{http://arxiv.org/abs/2012.08347}.
\newblock arXiv:2012.08347 [cs].

\bibitem[Wang et~al.(2023)Wang, Dong, Cheng, Liu, Yan, Gao, and Wei]{wang_augmenting_2023}
Weizhi Wang, Li~Dong, Hao Cheng, Xiaodong Liu, Xifeng Yan, Jianfeng Gao, and Furu Wei.
\newblock Augmenting {Language} {Models} with {Long}-{Term} {Memory}.
\newblock \emph{Advances in Neural Information Processing Systems}, 36:\penalty0 74530--74543, December 2023.
\newblock URL \url{https://proceedings.neurips.cc/paper_files/paper/2023/hash/ebd82705f44793b6f9ade5a669d0f0bf-Abstract-Conference.html}.

\bibitem[Wang et~al.(2021)Wang, Byrnes, Wang, Sun, Ma, Chen, Wu, and Xue]{wang_data_2021}
Zihan Wang, Olivia Byrnes, Hu~Wang, Ruoxi Sun, Congbo Ma, Huaming Chen, Qi~Wu, and Minhui Xue.
\newblock Data {Hiding} with {Deep} {Learning}: {A} {Survey} {Unifying} {Digital} {Watermarking} and {Steganography}, July 2021.
\newblock URL \url{https://arxiv.org/abs/2107.09287v3}.

\bibitem[Wei et~al.(2023)Wei, Wang, Schuurmans, Bosma, Ichter, Xia, Chi, Le, and Zhou]{wei_chain--thought_2023}
Jason Wei, Xuezhi Wang, Dale Schuurmans, Maarten Bosma, Brian Ichter, Fei Xia, Ed~Chi, Quoc Le, and Denny Zhou.
\newblock Chain-of-{Thought} {Prompting} {Elicits} {Reasoning} in {Large} {Language} {Models}, January 2023.
\newblock URL \url{http://arxiv.org/abs/2201.11903}.
\newblock arXiv:2201.11903 [cs].

\bibitem[Wei \& Heim(2024)Wei and Heim]{wei_designing_2024}
Kevin Wei and Lennart Heim.
\newblock Designing {Incident} {Reporting} {Systems} for {Harms} from {AI}.
\newblock May 2024.

\bibitem[Weidinger et~al.(2023)Weidinger, Rauh, Marchal, Manzini, Hendricks, Mateos-Garcia, Bergman, Kay, Griffin, Bariach, Gabriel, Rieser, and Isaac]{weidinger_sociotechnical_2023}
Laura Weidinger, Maribeth Rauh, Nahema Marchal, Arianna Manzini, Lisa~Anne Hendricks, Juan Mateos-Garcia, Stevie Bergman, Jackie Kay, Conor Griffin, Ben Bariach, Iason Gabriel, Verena Rieser, and William Isaac.
\newblock Sociotechnical {Safety} {Evaluation} of {Generative} {AI} {Systems}, October 2023.
\newblock URL \url{http://arxiv.org/abs/2310.11986}.
\newblock arXiv:2310.11986 [cs].

\bibitem[Wills(2024)]{wills_care_2024}
Peter Wills.
\newblock Care for {Chatbots}, May 2024.
\newblock URL \url{https://papers.ssrn.com/abstract=4814272}.

\bibitem[Wolf et~al.(2020)Wolf, Debut, Sanh, Chaumond, Delangue, Moi, Cistac, Rault, Louf, Funtowicz, Davison, Shleifer, von Platen, Ma, Jernite, Plu, Xu, Le~Scao, Gugger, Drame, Lhoest, and Rush]{wolf_transformers_2020}
Thomas Wolf, Lysandre Debut, Victor Sanh, Julien Chaumond, Clement Delangue, Anthony Moi, Pierric Cistac, Tim Rault, Remi Louf, Morgan Funtowicz, Joe Davison, Sam Shleifer, Patrick von Platen, Clara Ma, Yacine Jernite, Julien Plu, Canwen Xu, Teven Le~Scao, Sylvain Gugger, Mariama Drame, Quentin Lhoest, and Alexander Rush.
\newblock Transformers: {State}-of-the-{Art} {Natural} {Language} {Processing}.
\newblock In Qun Liu and David Schlangen (eds.), \emph{Proceedings of the 2020 {Conference} on {Empirical} {Methods} in {Natural} {Language} {Processing}: {System} {Demonstrations}}, pp.\  38--45, Online, October 2020. Association for Computational Linguistics.
\newblock \doi{10.18653/v1/2020.emnlp-demos.6}.
\newblock URL \url{https://aclanthology.org/2020.emnlp-demos.6}.

\bibitem[Wu et~al.(2023)Wu, Bansal, Zhang, Wu, Li, Zhu, Jiang, Zhang, Zhang, Liu, Awadallah, White, Burger, and Wang]{wu_autogen_2023}
Qingyun Wu, Gagan Bansal, Jieyu Zhang, Yiran Wu, Beibin Li, Erkang Zhu, Li~Jiang, Xiaoyun Zhang, Shaokun Zhang, Jiale Liu, Ahmed~Hassan Awadallah, Ryen~W. White, Doug Burger, and Chi Wang.
\newblock {AutoGen}: {Enabling} {Next}-{Gen} {LLM} {Applications} via {Multi}-{Agent} {Conversation} {Framework}.
\newblock 2023.
\newblock \_eprint: 2308.08155.

\bibitem[Xie et~al.(2024)Xie, Zhang, Chen, Li, Zhao, Cao, Hua, Cheng, Shin, Lei, Liu, Xu, Zhou, Savarese, Xiong, Zhong, and Yu]{xie_osworld_2024}
Tianbao Xie, Danyang Zhang, Jixuan Chen, Xiaochuan Li, Siheng Zhao, Ruisheng Cao, Toh~Jing Hua, Zhoujun Cheng, Dongchan Shin, Fangyu Lei, Yitao Liu, Yiheng Xu, Shuyan Zhou, Silvio Savarese, Caiming Xiong, Victor Zhong, and Tao Yu.
\newblock {OSWorld}: {Benchmarking} {Multimodal} {Agents} for {Open}-{Ended} {Tasks} in {Real} {Computer} {Environments}, April 2024.
\newblock URL \url{http://arxiv.org/abs/2404.07972}.
\newblock arXiv:2404.07972 [cs].

\bibitem[Zhan et~al.(2024)Zhan, Liang, Ying, and Kang]{zhan_injecagent_2024}
Qiusi Zhan, Zhixiang Liang, Zifan Ying, and Daniel Kang.
\newblock {InjecAgent}: {Benchmarking} {Indirect} {Prompt} {Injections} in {Tool}-{Integrated} {Large} {Language} {Model} {Agents}, March 2024.
\newblock URL \url{http://arxiv.org/abs/2403.02691}.
\newblock arXiv:2403.02691 [cs].

\bibitem[Zhang et~al.(2023)Zhang, Edelman, Francati, Venturi, Ateniese, and Barak]{zhang_watermarks_2023}
Hanlin Zhang, Benjamin~L. Edelman, Danilo Francati, Daniele Venturi, Giuseppe Ateniese, and Boaz Barak.
\newblock Watermarks in the {Sand}: {Impossibility} of {Strong} {Watermarking} for {Generative} {Models}, November 2023.
\newblock URL \url{http://arxiv.org/abs/2311.04378}.
\newblock arXiv:2311.04378 [cs].

\bibitem[Zittrain(2024)]{zittrain_we_2024}
Jonathan Zittrain.
\newblock We {Need} to {Control} {AI} {Agents} {Now}, July 2024.
\newblock URL \url{https://www.theatlantic.com/technology/archive/2024/07/ai-agents-safety-risks/678864/}.
\newblock Section: Technology.

\bibitem[Ziv et~al.(2024)Ziv, {Lior Arzi}, {Eyal Paz}, {David Cross}, {Hiroki Suezawa}, {Naor Penso}, {Shai Sivan}, {Dineshwar Sahni}, {Maxim Kovalsky}, {Chenxi Wang}, {Roy Feintuch}, {Hadas Harel Lavie}, {Ronen Atias}, and {Gadi Evron}]{ziv_open_2024}
Neatsun Ziv, {Lior Arzi}, {Eyal Paz}, {David Cross}, {Hiroki Suezawa}, {Naor Penso}, {Shai Sivan}, {Dineshwar Sahni}, {Maxim Kovalsky}, {Chenxi Wang}, {Roy Feintuch}, {Hadas Harel Lavie}, {Ronen Atias}, and {Gadi Evron}.
\newblock Open {Software} {Supply} {Chain} {Attack} {Reference} ({OSC}\&{R}), 2024.
\newblock URL \url{https://pbom.dev/}.

\bibitem[Zou et~al.(2023)Zou, Wang, Carlini, Nasr, Kolter, and Fredrikson]{zou_universal_2023}
Andy Zou, Zifan Wang, Nicholas Carlini, Milad Nasr, J.~Zico Kolter, and Matt Fredrikson.
\newblock Universal and {Transferable} {Adversarial} {Attacks} on {Aligned} {Language} {Models}, July 2023.
\newblock URL \url{https://arxiv.org/abs/2307.15043v2}.

\end{thebibliography}


\clearpage

\appendix

\section*{Appendix}

\section{A More Detailed Definition of an Instance}\label{sec:detailed-definition-instances}
A \textbf{system} is a model (e.g., a set of parameters), along with software used to run the model and provide other user functions. For example, ChatGPT with GPT-4 as a backend comprises both the weights of (a particular version of) GPT-4 and the software used to facilitate chat interactions (as opposed to an ``autocomplete'' function, as in the OpenAI playground). 

\begin{figure*}[!htb]
\centering
    \includegraphics[width=0.6\textwidth]{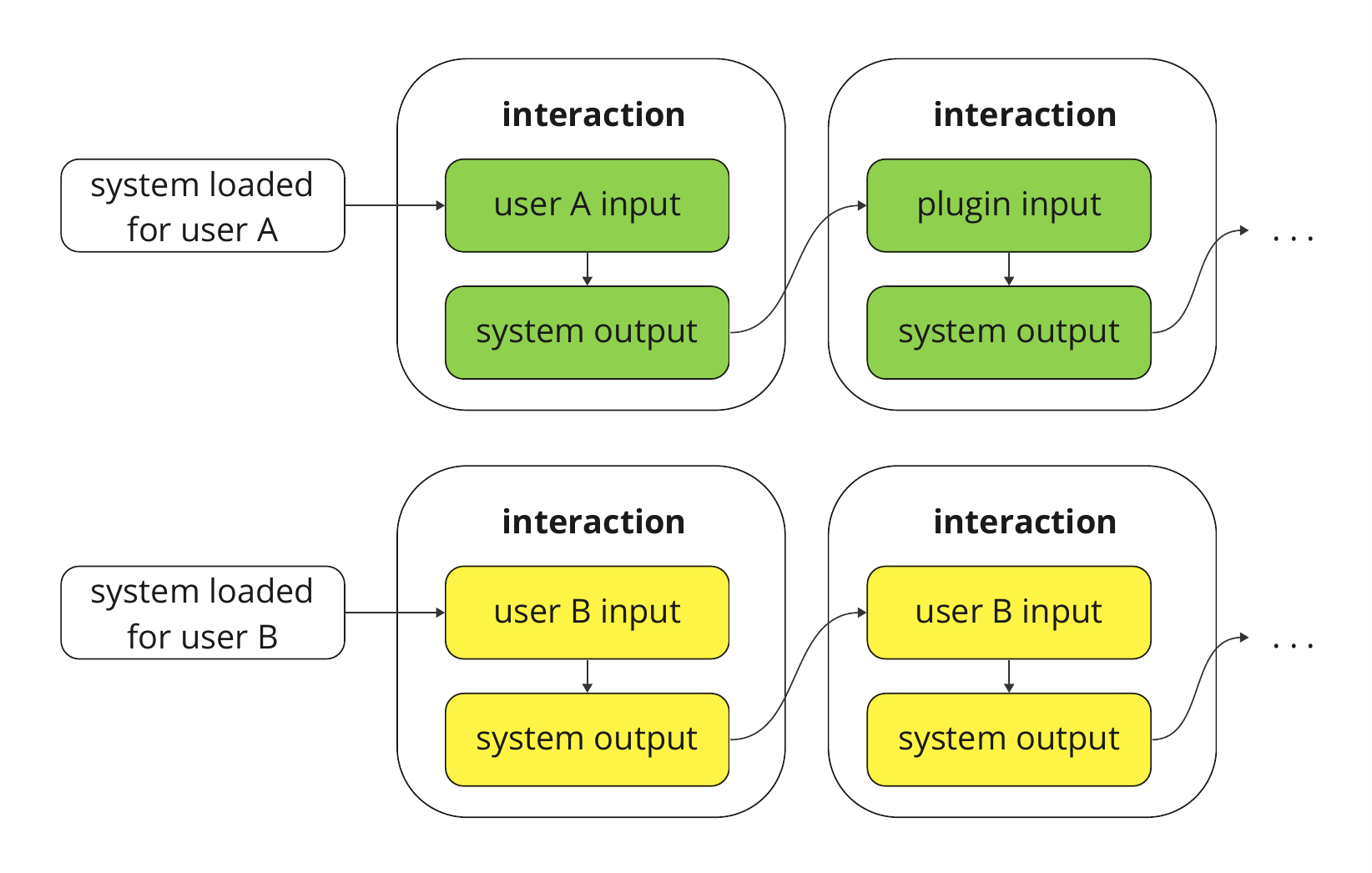}
    \caption{This figure depicts two users that use the same system (e.g., both use ChatGPT with the GPT-4 backend). System-specific information attached to an ID should be useful to both users. By \textbf{loaded}, we mean that the system is ready to accept inputs for the first time.}
    \label{fig:systems}
\end{figure*}

To understand the difference between a system and an instance, consider the information that would be useful to attach to a system-specific ID, so as to inform decisions about interacting with the system. Such information could not depend (too much) on a system's inputs, since different inputs might lead to different behavior. In other words, the information should ideally be valid regardless of a user's interaction history. Represented visually, the information would have to be valid for both top and bottom flow charts in \Cref{fig:systems}.

\begin{figure*}
\centering
    \includegraphics[width=0.7\textwidth]{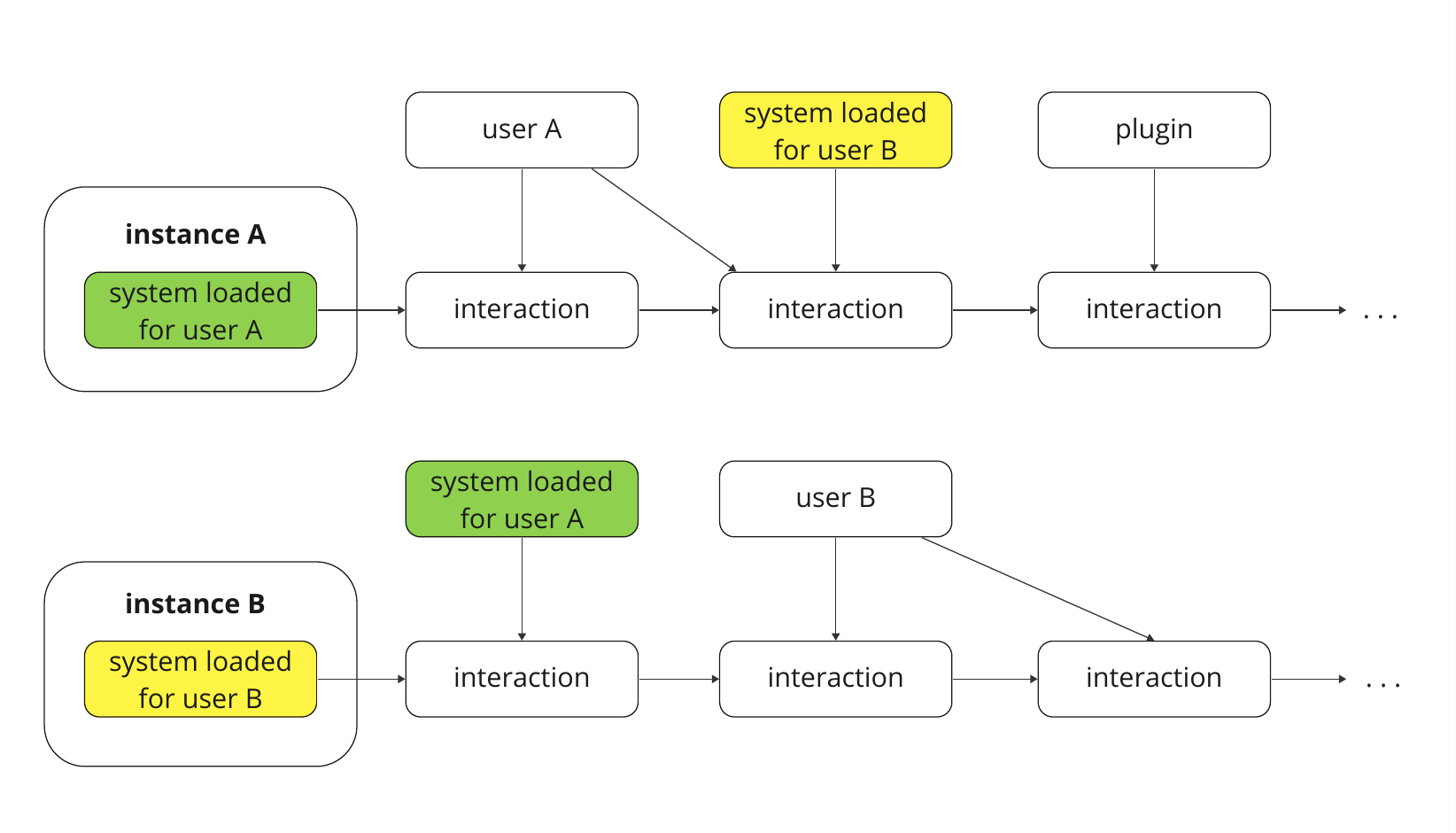}
    \caption{An \textbf{instance} is an abstraction that corresponds to a creation event, where a system is loaded for a user, and an interaction history. Instance-specific IDs could help to inform interaction decisions. For example, information about instances $A$'s earlier interactions (such as malfunctions) may be useful when instance $B$ interacts with instance $A$.}
    \label{fig:instances}
\end{figure*}

An \textbf{instance} is an abstraction that corresponds to a (initial) user (which could be a human, a group of humans, a software system, etc) and an interaction history. 
In \Cref{fig:instances}, we provide a visual depiction of two instances that interact with each other. 

We define instances in this way so that instance-specific IDs can take into account information that is causally relevant to a given interaction.\footnote{There may be information that is not causally relevant, but which may still help to predict the instance's behavior, such as model evaluations or the behaviour for other instances. It remains unclear how much of this information to attach to a given instance's ID.} For example, in the top half of \Cref{fig:instances}, information about instances $A$'s earlier interactions (such as malfunctions) may be useful when instance $B$ interacts with instance $A$. Actions taken by another instance are not, by default, causally relevant to the behaviour of another instance. Yet, instances can affect each other through direct interaction or on changing shared states of the world (e.g., using the same bank account).

\begin{figure*}
     \centering
     \begin{subfigure}[b]{\textwidth}
         \centering
         \includegraphics[width=0.7\textwidth]{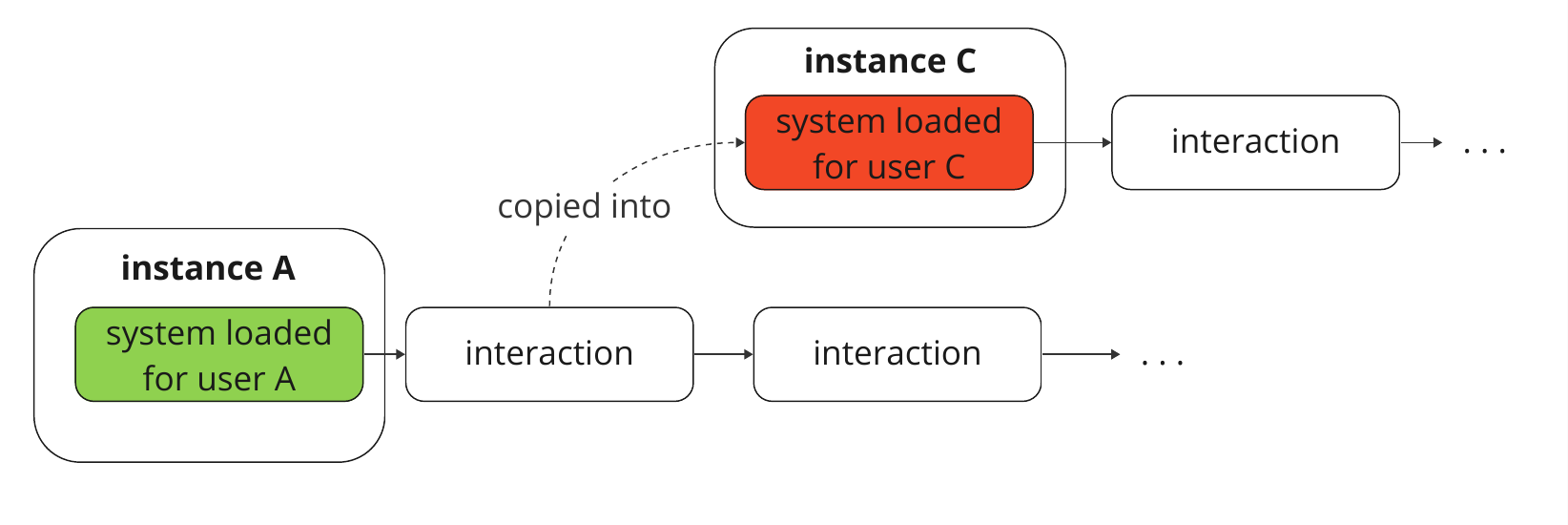}
         \caption{In this example of a branch, the inputs of instance $A$ are copied for user $C$ to create a new instance $C$.}
         \label{fig:branch}
     \end{subfigure}
     
     \begin{subfigure}[b]{\textwidth}
         \centering
         \includegraphics[width=0.7\textwidth]{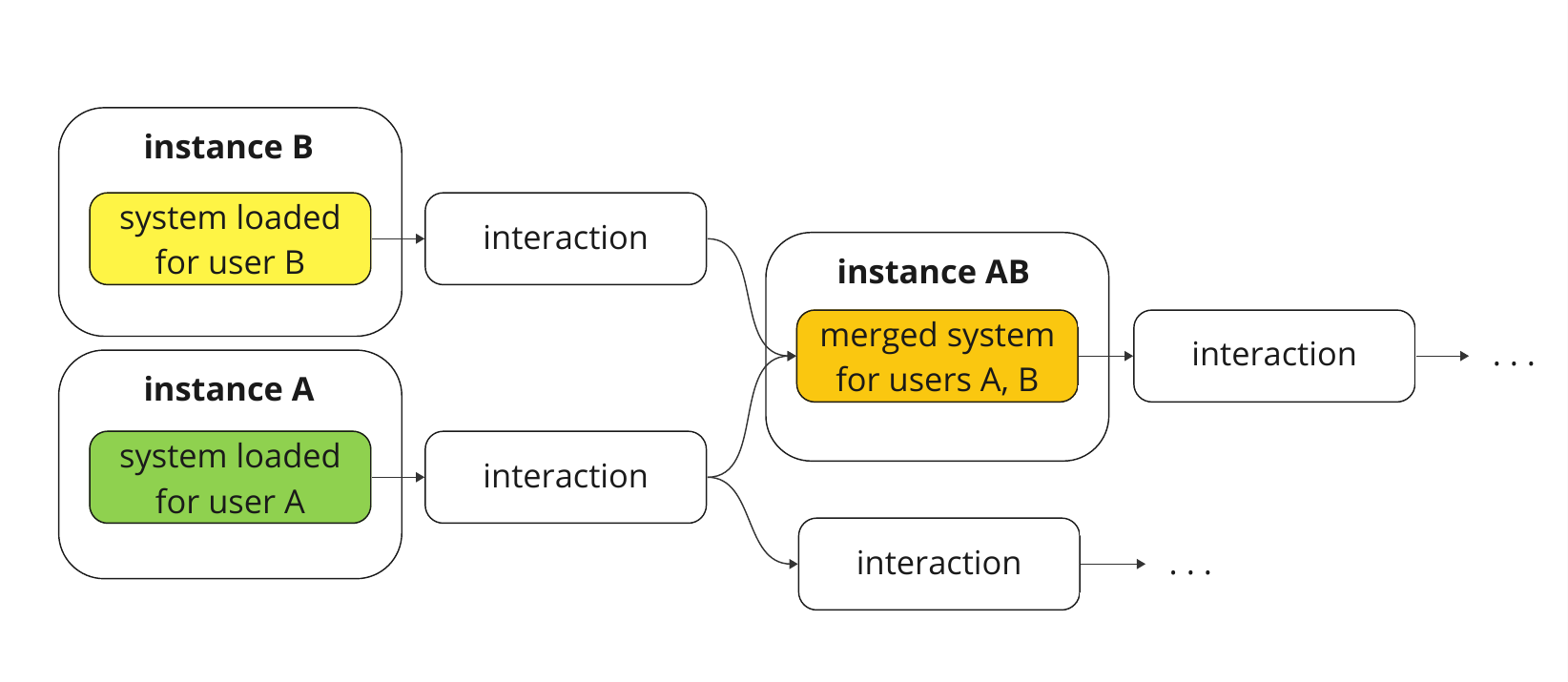}
         \caption{In this example of a merge, the instances $A$ and $B$ are merged into a new instance $AB$, which has access to (potentially a subset of) the inputs of $A$ and $B$.}
         \label{fig:merge}
     \end{subfigure}
     \caption{We illustrate how we define instances in the event that they are copied or combined.}
     \label{fig:edge-cases}
\end{figure*}

Since AI systems can be copied and combined, there are some additional edge cases for how to define instances. In a \textbf{branch} (see \Cref{fig:branch}), 
Two instances share a past context. 
For example, a regeneration of a response (possible in ChatGPT, for one) creates two branches. Since branches can behave independently, we suggest treating branches as separate instances. In a \textbf{merge} (see \Cref{fig:merge}), two instances come together to form a new system. For example, suppose software $S$ accepts inputs from separate users, engages the instances of two users in a debate based on the inputs, and finally outputs a result to both users. Users $A$ and $B$ could use $S$ to create a new instance $AB$, based on their separate instances. As with branching, we treat $AB$ as a separate instance because the users could continue running their original instances in tandem with $AB$.

\begin{figure*}
     \centering
     \begin{subfigure}[b]{\textwidth}
         \centering
         \includegraphics[width=0.7\textwidth]{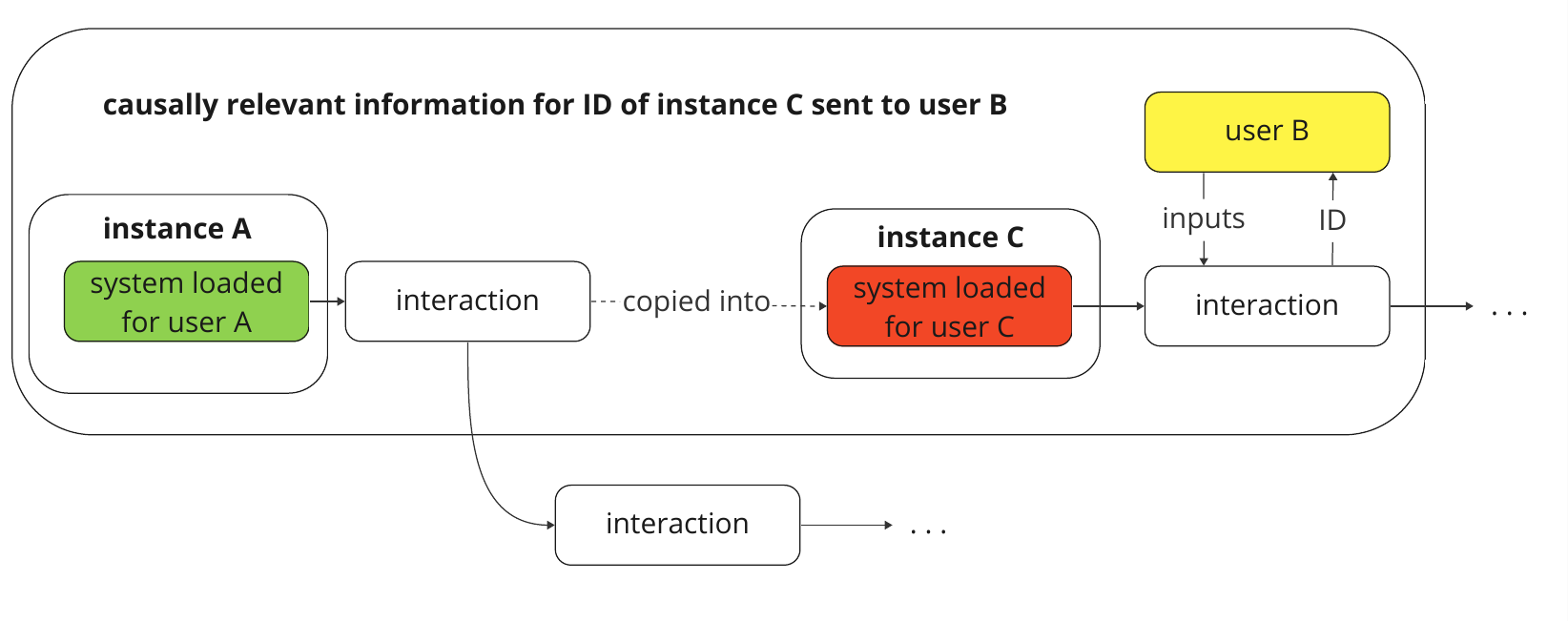}
         \caption{In the event of a branch, a user $B$ that interacts with instance $C$ may find information about instance $A$ (e.g., instance $A$ was involved in an incident) to be useful.}
         \label{fig:branch-id}
     \end{subfigure}
     
     \begin{subfigure}[b]{\textwidth}
         \centering
         \includegraphics[width=0.7\textwidth]{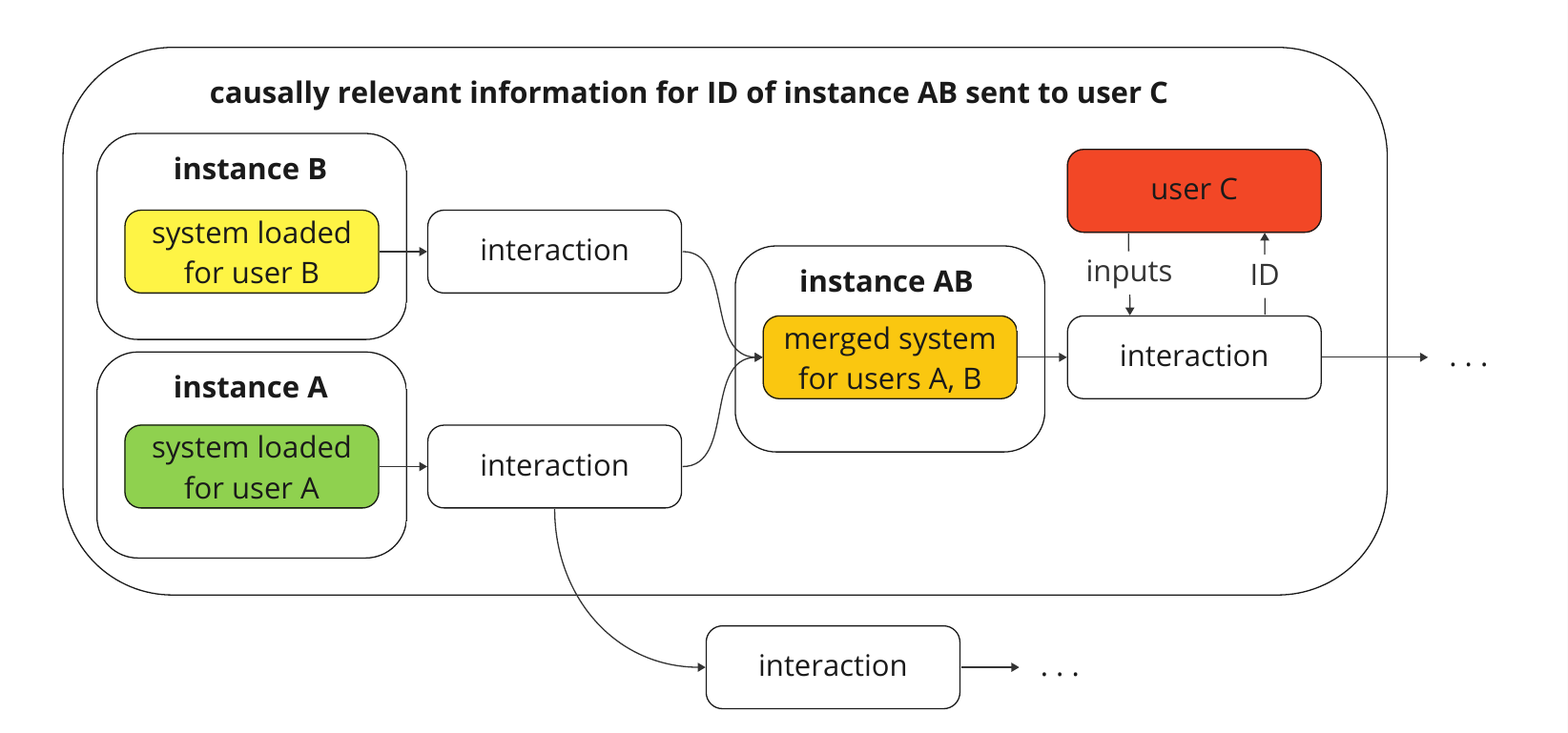}
         \caption{In the event of a merge, a user $C$ that interacts with instance $AB$ may find information (e.g., instance $B$ is not robust) about either instances $A$ or $B$ to be helpful.}
         \label{fig:merge-id}
     \end{subfigure}
     \caption{We visually represent the information that an ID for an instance could take into account.}
     \label{fig:id-cases}
\end{figure*}

If an instance is the result of a branch or a merge, it may be useful for the instance's ID to contain information about ancestor instances. In \Cref{fig:branch-id}, since the context of instance $A$ is copied into instance $C$, any incidents associated with $A$ may be helpful for user $B$'s interaction decisions. Similarly, in \Cref{fig:merge-id}, information about instances $A$ and $B$ may be relevant for user $C$'s decision about whether or not to trust the outputs of instance $AB$. Nonetheless, information from ancestor instances could be excluded for a variety of reasons, such as user privacy. Furthermore, it could be possible for information from descendant instances, or instances in another branch, to be relevant for the behavior of a given instance. For example, incidents with user $B$ from time $t + 1$ may be informative for user $A$ that interacted with an instance at time $t$. We leave further analysis of how to delimit the sources of ID information to future work.\footnote{Links to ancestor IDs make other potential definitions of an instance equivalent, from the perspective of the functions of an ID. For example, in \Cref{fig:branch} we could have defined instance $C$ as encompassing interaction history up to instance $A$. This definition would not change the causally relevant information with respect to instance $C$ in \Cref{fig:branch-id}. }

As we have defined it, an instance is causally independent of other instances unless it interacts with them, such as by processing their outputs. This causal independence is crucial to IDs, as we discussed above. Yet, the abstraction of an instance could be leaky depending on system implementation details. For example, if a single GPU handles the operation of two separate instances, implementation errors or optimization strategies (e.g., approximate matrix multiplication) could potentially result in computational interference between the instances.

\section{Maintaining Access to IDs}\label{sec:maintaining-accessibility}
Suppose a primary party $P$ receives an output and ID from an AI system. 
If it is desirable for secondary parties to access the ID, $P$ may have additional responsibilities. Some potential situations follow. 

\textbf{Chains of deployers}: There may be multiple deployers involved in the operation of an AI system. For example, deployer $A$ could provide an API for system $X$ to customer $B$, who creates system $Y$ and serves $Y$ to user $C$. To ensure that $C$ can access the $X$'s ID, $B$ should link to the ID provided by $A$, or should include the information in the ID of $Y$. 

\textbf{Service providers}: If a service provider receives an ID, it should ensure that those who observe or are affected by AI system's actions can also see the ID. For instance, a bank that facilitates a financial transaction from an AI system should make the latter's ID known to the other party of the transaction.

The provider may have to work with other parties to maintain visibility of the ID. As an example, consider a service that allows an AI system to post on a social media platform. The provider may have written the software that allows the AI system to interact with the platform's API, but the social media company develops the API and manages the platform. While the provider could include the AI system's ID as text within the post, doing so may be infeasible because of text limits or would otherwise be obstructive to the user experience. Rather, the provider should work with the social media company to include the ID as metadata, readily accessible to the user through a visible icon. 

\textbf{Users}: Users could receive outputs with attached IDs, whether as a watermark or metadata. 
To counter inadvertent removal of IDs, deployers could add labels to inform users. For example, Facebook and Instagram automatically add labels indicating AI origin when a user shares a photo generated with the Meta AI feature \citep{bickert_our_2024}. To counter intentional removal of IDs, better watermarking techniques may be required. Yet, it remains unclear how effective watermarking may ultimately be \citep{zhang_watermarks_2023,gleichauf_digital_2024}.

\end{document}